\documentclass{article}

\PassOptionsToPackage{numbers, compress}{natbib}
\usepackage[final]{neurips_2021}
\usepackage{microtype}
\usepackage{graphicx}
\usepackage{subfigure}
\usepackage{booktabs} % for professional tables

\usepackage{amsmath,amsfonts,bm}

\def\eqref#1{equation~\ref{#1}}
\def\1{\bm{1}}

\DeclareMathAlphabet{\mathsfit}{\encodingdefault}{\sfdefault}{m}{sl}
\SetMathAlphabet{\mathsfit}{bold}{\encodingdefault}{\sfdefault}{bx}{n}

\usepackage{pifont}
\usepackage{amssymb}
\usepackage{autobreak}
\usepackage{amsmath}
\usepackage{adjustbox}
\usepackage{longtable}
\usepackage{bm,nicefrac}
\usepackage{mathrsfs}
\usepackage{amsthm}
\theoremstyle{plain}

\usepackage{multirow}
\def\*#1{\mathbf{#1}}
\def\*#1{\mathbf{#1}}
\usepackage{graphicx}
\usepackage{subfigure}
\usepackage[normalem]{ulem}
\usepackage{mathtools}
\usepackage{wrapfig}

\def\*#1{\mathbf{#1}}
\usepackage{color, colortbl}
\definecolor{Gray}{gray}{0.9}
\usepackage{xcolor}

\newcommand{\Din}{\mathcal{D}_\mathrm{in}}
\newcommand{\Dout}{\mathcal{D}_\mathrm{out}}

\newcommand{\calP}{\mathcal{P}}

\newcommand{\calX}{\mathcal{X}}
\newcommand{\calY}{\mathcal{Y}}

\newcommand{\bx}{\mathbf{x}}

\usepackage{hyperref}

\title{Can multi-label classification networks 
know what they don't know?}

\author{%
  Haoran Wang\thanks{Equal contribution. Work done while H.W was working at UW-Madison as an undergraduate researcher.} \\
  Information Networking Institute\\
 Carnegie Mellon University\\
  \texttt{haoranwa@andrew.cmu.edu} \\
  \And 
 Weitang Liu\\
Department of Computer Science and Eng.\\
University of California, San Diego\\
\texttt{wel022@ucsd.edu} \\
\And
Alex Bocchieri\\
Department of Computer Sciences\\
University of Wisconsin-Madison\\
\texttt{abocchieri@wisc.edu}
\And
Yixuan Li$^*$\\
Department of Computer Sciences\\
University of Wisconsin-Madison\\
\texttt{sharonli@cs.wisc.edu}

}

\begin{document}
\maketitle

\begin{abstract}
% Out-of-distribution (OOD) detection is essential to prevent anomalous inputs from causing a model to fail during deployment. Improved methods for OOD detection in multi-class classification have emerged, while OOD detection methods for multi-label classification remain limited and use rudimentary techniques. 
% %In this paper, we propose an energy-based method for OOD detection in the multi-label setting, which estimates the OOD indicator scores jointly from multiple labels. 
% We propose \emph{JointEnergy}, a simple and effective method, which estimates the OOD indicator scores by aggregating energy scores from individually independent labels. We show that \emph{JointEnergy} can be mathematically interpreted from a joint likelihood perspective. Our results show consistent improvement over previous methods that are based on the maximum-valued scores, which fail to capture information from other possible labels. We demonstrate the effectiveness  on three common multi-label classification benchmarks including MS-COCO, PASCAL-VOC, and NUS-WIDE. We show that \emph{JointEnergy} reduces the FPR95 by up to 7.8\% compared to the previous best baseline, establishing state-of-the-art performance. 

Estimating out-of-distribution (OOD) uncertainty is a major challenge for safely deploying machine learning models in the open-world environment. Improved methods for OOD detection in multi-class classification have emerged, while OOD detection methods for multi-label classification remain underexplored and use rudimentary techniques. We propose \emph{JointEnergy}, a simple and effective method, which estimates the OOD indicator scores by aggregating label-wise energy scores from multiple labels. We show that {JointEnergy}  can be mathematically interpreted from a joint likelihood perspective. Our results show consistent improvement over previous methods that are based on the maximum-valued scores, which fail to capture joint information from multiple labels. We demonstrate the effectiveness of our method on three common multi-label classification benchmarks, including MS-COCO, PASCAL-VOC, and NUS-WIDE. We show that {JointEnergy} can reduce the FPR95 by up to {10.05}\% compared to the previous best baseline, establishing \emph{state-of-the-art} performance.

\end{abstract}
\section{Introduction}
%Out-of-distribution (OOD) detection is central for reliably deploying machine learning models in open-world environments, where new forms of test-time data may appear that were nonexistent during the training time. 
Despite many breakthroughs in machine learning, formidable obstacles obstruct its deployment in the real world, where a model can encounter unknown out-of-distribution (OOD) samples. %This gives rise to the importance of {out-of-distribution (OOD) detection}, which allows the
%learner to identify and handle OOD inputs safely.
The problem of OOD detection has gained significant research attention lately~\citep{chen2021robustifying, hsu2020generalized, huang2021mos, 
lakshminarayanan2017simple,
lee2018simple,
liang2018enhancing, liu2020energy, lin2021mood, sun2021tone}. OOD detection aims to identify test-time inputs that have no label intersection with training classes, thus should not be predicted by the model. Previous studies have primarily focused on detecting OOD examples in multi-class classification, where each sample is assigned to a single label.  Unfortunately, this can be unrealistic in many real-world applications where images often have \emph{multiple labels} of interest. 
% \SL{Need some convincing example. For example, perception model in self-driving needs to classify multiple objects in the scene. And then maybe use this as running example for Figure 1 and rest of paper. This example ideally should make readers want to immediately read on! Caveat of using self-driving car example is that we don't have evaluation to support this application..} 
For example, 
%self-driving cars must differentiate between the road, traffic signs, and obstacles within a frame. 
in medical imaging, multiple abnormalities may be present in a medical scan~\citep{wang2017chestx}. %Multi-label classification is desirable since there is no constraint on the number of classes an instance can be assigned to.
%A machine learning model's reliability is determined by how well it generalizes to unseen data. 
%During the development phase, a high validation accuracy indicates the model was trained successfully and is ready for real-world deployment. 
%A model may fail drastically at its task 
%if a test-time input is significantly different from the training data, which may arise when the training data is not completely representative of all possible real-world data, or
%when new forms of data appear during the deployment phase that were nonexistent during the training phase. 

% Recent studies have mostly focused on detecting OOD examples in a multi-class (single-label) classification task \textcolor{red}{(cite papers?)}. Deep learning models for multi-label classification have been extensively developed and have many applications \textcolor{red}{(cite papers)} \citep{wang2016cnn}. Multi-label classification is often better suited than multi-class classification in the image domain since images can have multiple objects, actions, and attributes of interest.

\begin{figure*}[t]
\begin{center}
\includegraphics[width=137mm]{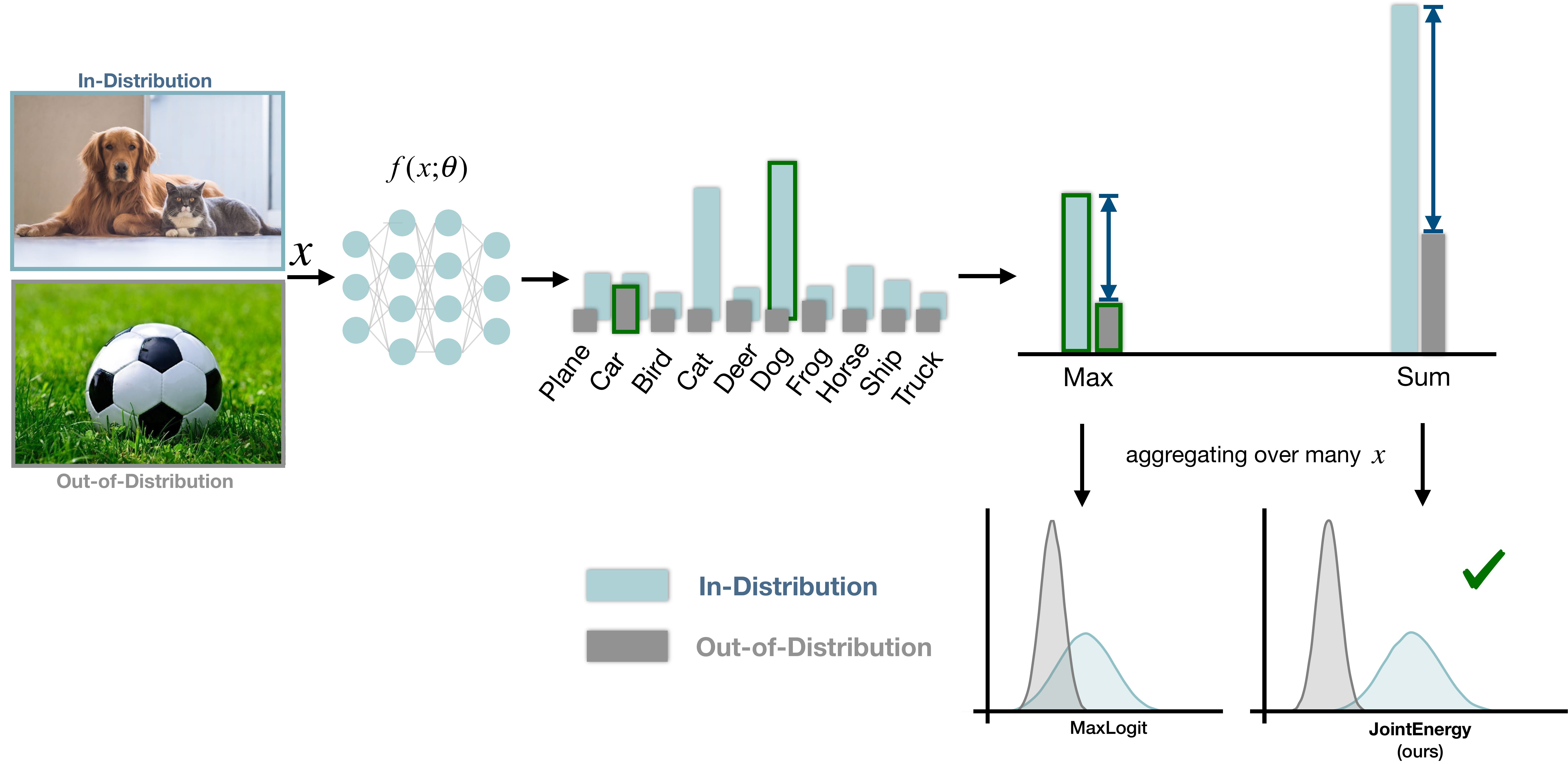}
\vspace{-0.3cm}
\caption{\footnotesize Out-of-distribution detection for multi-label classification networks. During inference time, input $\*x$ is passed through classifier $f$, and label-wise scores are computed for each label. OOD indicator scores are either the maximum-valued score (denoted by green outlines) or the sum of all scores. Taking the sum results in a larger difference in scores and more separation between in-distribution and OOD inputs (denoted by red lines), resulting in better OOD detection. Plots in the bottom right depict the probability densities of MaxLogit~\cite{hendrycks2019benchmark}  versus \emph{JointEnergy} (ours).
% Energy-based out-of-distribution detection for multi-label classification. The energy can be used as a scoring function for any pre-trained neural network. During inference time, for a given input $\*x$, the class-wise energy score $E_y(\*x)$ is calculated for each label, which will be aggregated across all labels. The OOD detector will classify the input as OOD if the aggregated energy score is smaller than the threshold value. 
%The threshold in practice can be chosen on the in-distribution data so that true positive rate is 95\%.
\vspace{-0.5cm}
\label{fig:banner}}
\end{center}
\end{figure*}

Currently, a critical research gap exists in developing and evaluating OOD detection algorithms for multi-label classification tasks that are more applicable to the real world. While one may expect solutions for multi-class setting should transfer to the multi-label setting, we show that this is far from the truth. 
%OOD detection in multi-label classification remains relatively underexplored. 
The main challenges posed in  multi-label setting stem from the need to estimate uncertainty by \emph{jointly leveraging the information across different labels,} as opposed to relying on one dominant label. Our analysis reveals that simply using the largest model output (\emph{i.e.}, MaxLogit) can be {limiting}. As a simple illustration, we contrast in Figure~\ref{fig:banner} of estimating OOD uncertainty using joint vs. single label information. MaxLogit can only capture the difference between the dominant outputs for \texttt{dog} (in-distribution) and \texttt{car} (OOD), while positive information from another dominant label \texttt{cat} (in-distribution) is dismissed.

In this paper, we address the important problem of OOD uncertainty estimation in the multi-label classification setting, and propose a simple and surprisingly effective method that jointly characterizes uncertainty from multiple labels. As a major advantage, our method circumvents the challenge to {directly} estimate the joint likelihood using generative models, which  can be can computationally intractable to train and optimize, especially on multi-label datasets~\cite{hinz2019generating}. 
%Our key idea is to combine the label-wise energy scores derived from individually independent classes, and produce a joint energy estimation that is lower for in-distribution data and higher for OOD data. %While the joint estimation across labels can be exponentially more difficult compared to the multi-class setting (single label), we show that aggregating energies can be surprisingly convenient and computationally tractable. \SL{haoran please double check if this makes sense...we want to say something that directly estimating the joint likelihood can be difficult} %Intuitively, an input with multiple dominant labels is more likely to be in-distribution
%, which is the key aspect that \emph{JointEnergy} capitalizes on. 
Formally, our proposed method, \emph{JointEnergy}, derives a novel OOD score by combining label-wise energies over all labels. 
Despite its simplicity, we show that the {JointEnergy}  can be theoretically interpreted from a joint likelihood perspective. The joint likelihood allows separability between in-distribution vs. OOD data, since OOD data is expected to have lower joint likelihood (\emph{i.e.}, not associated with any of the labels). In contrast, having multiple dominant labels is indicative of an in-distribution input, which is the key aspect that {JointEnergy} captures. As shown in Figure~\ref{fig:banner}, {JointEnergy} effectively amplifies the difference in scores between in-distribution and OOD inputs, compared to MaxLogit. %Below we describe our contributions in  detail.%\SL{Quick explanation on what conditional sum means?} Our method is parameter-free and can be conveniently used for any pre-trained multi-label classification model. 

  %First, we propose a theoretically motivated scoring function, \emph{JointEnergy}, that is based on the aggregation over label-wise energy scores. 
  
  Extensive experiments show that {JointEnergy}  outperforms existing methods on three common multi-label classification tasks, establishing {state-of-the-art} performance. For example, on a DenseNet trained with MS-COCO~\citep{lin2014microsoft}, our method reduces the false positive rate (at 95\% TPR) by {10.05}\% when evaluated against OOD data from ImageNet~\cite{deng2009imagenet}, compared to the best performing baselines. 
  %Consistent performance improvement is also observed on a different OOD test dataset Texture~\citep{cimpoi14describing}, 
  Consistent performance improvement is observed on other multi-label tasks including PASCAL-VOC~\citep{Everingham15} and NUS-WIDE~\citep{nus-wide-civr09}, as well as alternative network architecture.%Since \emph{JointEnergy} is a parameter-free measure, it is practically appealing and easy to use.

%\textcolor{red}{Our findings show that taking the maximum of a score limits all methods alike, including \emph{MaxEnergy}, and that summing labels' scores using previous methods is inferior to summing labels' energies, emphasizing the need for \emph{JointEnergy}. (Maybe this phrase should be moved to a discussion/conclusion section? This phrase might be redundant, next paragraph states something similar) (Wei: I made some changes in the next paragraph that emphasizes the difference between the ablasions studies. Maybe that will solve the issue?)} 
% Additionally, we perform a comparative analysis of how an alternative aggregation method affects OOD detection performance. In particular, we consider \emph{MaxEnergy}, which takes the maximum energy score among the individual labels as the OOD indicator score.
% On a DenseNet trained with the PASCAL-VOC dataset, \emph{MaxEnergy} yields 4.05\% lower FPR95 compared to \emph{JointEnergy}, which  underlines the importance of taking into account scores derived from multiple labels. 

%We further considered \emph{ExpJointEnergy}, which combines the exponential of negative label-wise energy scores and is theoretically connected to the input's probability density under certain distributional assumptions. Experiments show that \emph{ExpJointEnergy} outperforms prior baselines, and compares favorably with \emph{JointEnergy}. 

%nlike ODIN~\citeauthor{liang2017enhancing} or Mahalanobis~\citeauthor{lee2018simple} 

Importantly, our analysis 
%underlines the importance of properly choosing both the label-wise scoring function and the aggregation method. 
demonstrates a strong compatibility between the label-wise energy function and aggregation function, supported by both mathematical interpretation and empirical results.
 %demonstrates that energy scores are more compatible with the proposed summation aggregation method, compared to previous OOD scoring functions such as logit~\citep{hendrycks2019benchmark}, MSP~\citep{hendrycks2016baseline}, ODIN~\citep{liang2018enhancing}, and Mahalanobis distance~\citep{lee2018simple}.
%. $\rightarrow$ other than alternative aggregation methods, we also demonstrate that previous scoring functions, such as logit~\citep{hendrycks2019benchmark}, MSP~\citep{hendrycks2016baseline}, ODIN~\citep{liang2018enhancing}, and Mahalanobis distance~\citep{lee2018simple}, are less compatible with the summation aggregation method than our energy score.}
As an ablation, we explore the effectiveness of applying summation to popular OOD scoring functions~\cite{hendrycks2019benchmark,hendrycks2016baseline, liang2018enhancing, lee2018simple}. 
%, including logit~\citep{hendrycks2019benchmark}, softmax score~\citep{hendrycks2016baseline}, ODIN score~\citep{liang2018enhancing}, and Mahalanobis distance~\citep{lee2018simple}.
We find that summing labels' scores using previous methods is inferior to summing labels' energies, emphasizing the need for {JointEnergy}. For example, simply summing over the logits across labels results in up to 51.93\% degradation in FPR95 on MS-COCO. %In contrast, \emph{JointEnergy} does not suffer from this issue because the signs of label-wise energy scores are uniform.
%\textcolor{red}{(energy score $E_y(x)$ is negative, but we negate it when aggregating to make it positive, not sure if we should mention this detail here, might be fine as it is. In fact, this detail is mentioned after Eqn 7.)}.
%More importantly, label-wise energy is provably aligned with the probability density of the corresponding label's training data. 
Our study therefore underlines the importance of properly choosing both the label-wise scoring function and the aggregation method. 
%We show strong compatibility between the label-wise energy function and aggregation function, supported by both mathematical interpretation and empirical results. %The appealing property of energy score makes it suitable for multi-label OOD detection problem. 
Below we summarize our {key results and contributions}:
\begin{itemize}
    \item We propose a novel method \emph{JointEnergy}---addressing an important yet underexplored problem---OOD detection for multi-label classification networks. Our method establishes state-of-the-art performance, reducing the average FPR95 by up to {10.05}\%. We show theoretical interpretation, underpinning our method from a joint likelihood perspective.
    \item We conduct extensive ablations which reveals important insights for multi-label OOD uncertainty estimation under (1) different aggregation functions, (2) different label-wise OOD scoring functions, and (3) the compatibility thereof. 
    \item We curate three evaluation tasks in the multi-label setting from three real-world high-resolution image databases, which enables future research to evaluate OOD detection in a multi-label setting. Our code and dataset is released for reproducible research\footnote{Code and data is available:  \url{https://github.com/deeplearning-wisc/multi-label-ood}}.
\end{itemize}
\section{Background}

\textbf{Multi-label Classification} Multi-label classification is the supervised learning problem
where an instance may be associated with multiple labels. 
% The most well-known approach to multi-label classification
% is to simply train an independent classifier for each label.
% This is usually known in the literature as the binary relevance
% (BR) transformation~\cite{}.  
Let $\calX$ (resp. $\calY$) be the input (resp. output) space and let $\calP$ be a distribution over $\calX \times \calY$, and let $f : \calX \rightarrow \mathbb{R}^{|\calY|}$ be a neural network trained on samples drawn from $\calP$. An input can be associated with a subset of labels in $\calY=\{1,2,...,K\}$. This set is represented by a vector $\*y=[y_1,y_2,...,y_K]$, where $y_i=1$ if and only if label $i$ is associated with instance $\*x$, and 0 otherwise. We use a convolutional neural network with shared feature space and derive the multi-label output prediction. In contrast to learning completely disjoint classifiers~\cite{tsoumakas2007multi}, the end-to-end training with a shared feature space is computationally more efficient than training $K$ completely independent models. This has become a  de facto training mechanism for multi-label classification, with various domain applications~\cite{liu2017deep, liu2015multi, trohidis2008multi,  wang2017chestx}.
% The most well-known approach to multi-label classification is to simply train an independent classifier for each label.
% This is usually known in the literature as the binary relevance
% (BR) transformation~\cite{}. However, this method is limited this since
% dependencies between labels are not modelled. To better capture the dependencies among labels, we use a convolutional neural network with shared feature space, and derive the multi-label output prediction. Previous work employed
% a deep belief network for multi-label classification, and showed improvement over BR by taking into account the dependencies in the label space.

\textbf{Out-of-distribution Detection} The problem of OOD detection for multi-label classification is defined as follows. Denote by $\Din$ the marginal distribution of $\calP$ over $\calX$, which represents the distribution of in-distribution data. At test time, the environment can incur an out-of-distribution $\Dout$ over $\calX$. The goal of OOD detection is to define a decision function $G$ such that:
\begin{equation*}
    G(\bx; f) =
    \begin{cases}
    0 & \text{if } \bx \sim \Dout, \\
    1 & \text{if } \bx \sim \Din.
    \end{cases}
\end{equation*}
An input is considered an OOD if it does not contain any label in the in-distribution data. 
In practice, $\Dout$ is often defined by a distribution that simulates anomalies encountered during deployment time, such as samples from an irrelevant distribution whose label set has no intersection with $\calY$ and \emph{therefore should not be predicted by the model}.

\textbf{Energy Function} Liu \emph{et al.}~\cite{liu2020energy} first propose using free energy as a scoring function for OOD uncertainty estimation in the multi-class setting. Given a neural classifier $f(\*x):  \mathcal{X} \rightarrow \mathbb{R}^K$  that maps an input $\*x \in \mathcal{X}$ to K real-valued numbers as logits, %The energy model defines the probability distribution through logits $f_i(\*x)$ 
% We consider a multi-class neural classifier $f(\*x): \mathbb{R}^D \rightarrow \mathbb{R}^K$, which maps an input $\*x \in \mathbb{R}^D$ to K real-valued numbers as logits. 
a softmax function is used to derive a categorical distribution,
\begin{equation}\label{eq:softmax}
    p(y_i=1 \mid \*x) = \frac{e^{f_{y_i}(\*x)}}{\sum_{j=1}^K e^{f_{y_j}(\*x)}}.
\end{equation}
% which indicates the probability for an input $\*x$ to be of class $y_i$, with $i \in \{1,2,...,K\}$.  
% A multi-class classifier can be interpreted from an  energy-based perspective by viewing the logit $f_{y_i}(\*x)$ of class $y_i$ as an energy function $E(\*x,y_i) = -f_{y_i}(\*x)$.
The energy model defines the probability distribution through the logits. The transformation from logits to probability distribution is by the Boltzmann distribution:
%Therefore,  Equation~\ref{eq:softmax} can be rewritten as:
% \begin{align}
%     p(y_i \mid \*x) & = \frac{e^{-E(\*x,y_i)}}{\sum_{j=1}^K e^{-E(\*x,y_j)}} \\[5pt]
%     &= \frac{e^{-E(\*x,y_i)}}{e^{-E(\*x)}}.
% \end{align}
% By equalizing the two denominators above, we can express the \emph{free energy} function $E(\*x)$ for any given input $\*x\in \mathbb{R}^D$ in forms of:
% \begin{align}\label{eq:energy_softmax}
%   E(\*x)=- \text{log}\sum_{i=1}^K e^{f_{y_i}(\*x)}.
% \end{align}
% \cite{liu2020energy} propose to use free energy as a scoring function for OOD detection in the multi-class setting, with the theoretical motivation that free energy is provably aligned with the likelihood of data, \emph{i.e.,} $E(\*x) \propto -\log p(\*x)$.
% The energy model defines the probability distribution through the logits, which XXX(cite) proposes to use for out-of-distribution(OOD) detection. The translation from logits to probability distribution is by Gibbs Distribution:
\begin{align*}%\label{eq:gibbs}
    p(y_i=1 \mid \*x)  = \frac{e^{-E(\*x,y_i)}}{\int_{y'} e^{-E(\*x, y')}} 
    %  = \frac{e^{f_{y_i}(\*x)}}{\sum_{j=1}^K e^{f_{y_j}(\*x)}} \\ 
 = \frac{e^{-E(\*x,y_i)}}{e^{-E(\*x)}}.
\end{align*}
Therefore, a multi-class classifier can be interpreted from an energy-based perspective by viewing the logit $f_{y_i}(\*x)$ of class $y_i$ as an energy function $E(\*x,y_i) = -f_{y_i}(\*x)$.
By equalizing the two denominators above, the \emph{free energy} function $E(\*x)$ for any given input $\*x$ is:
\begin{align}\label{eq:energy_softmax}
  E(\*x)=- \text{log}\sum_{i=1}^K e^{f_{y_i}(\*x)}.
\end{align}

% where $E(x,y)$ is the logit for class $y$ with input $X$, which is image in our case, and $T$ is a temperature hyper-parameter.
% Because it is directly correspondent to data distribution, free energy function is proposed by XXX as an OOD detector:
% \begin{align}\label{eq:energy_softmax}
%   E(\*x;f)=- T\cdot \text{log}\sum_i^K e^{f_i(\*x)/T}.
% \end{align}
% where we define $E(\*x,y=i) = f_i(\*x)$. 

\section{Method}
\label{sec:method}

In this work, we propose a novel method for OOD detection in multi-label classification networks, where an input can have several labels (see Figure~\ref{fig:banner}). In what follows, we first introduce a label-wise energy function, and then propose \emph{JointEnergy} that can leverage the joint information across labels for OOD uncertainty estimation.

\begin{figure*}[t]
\begin{center}  
\includegraphics[width=135mm]{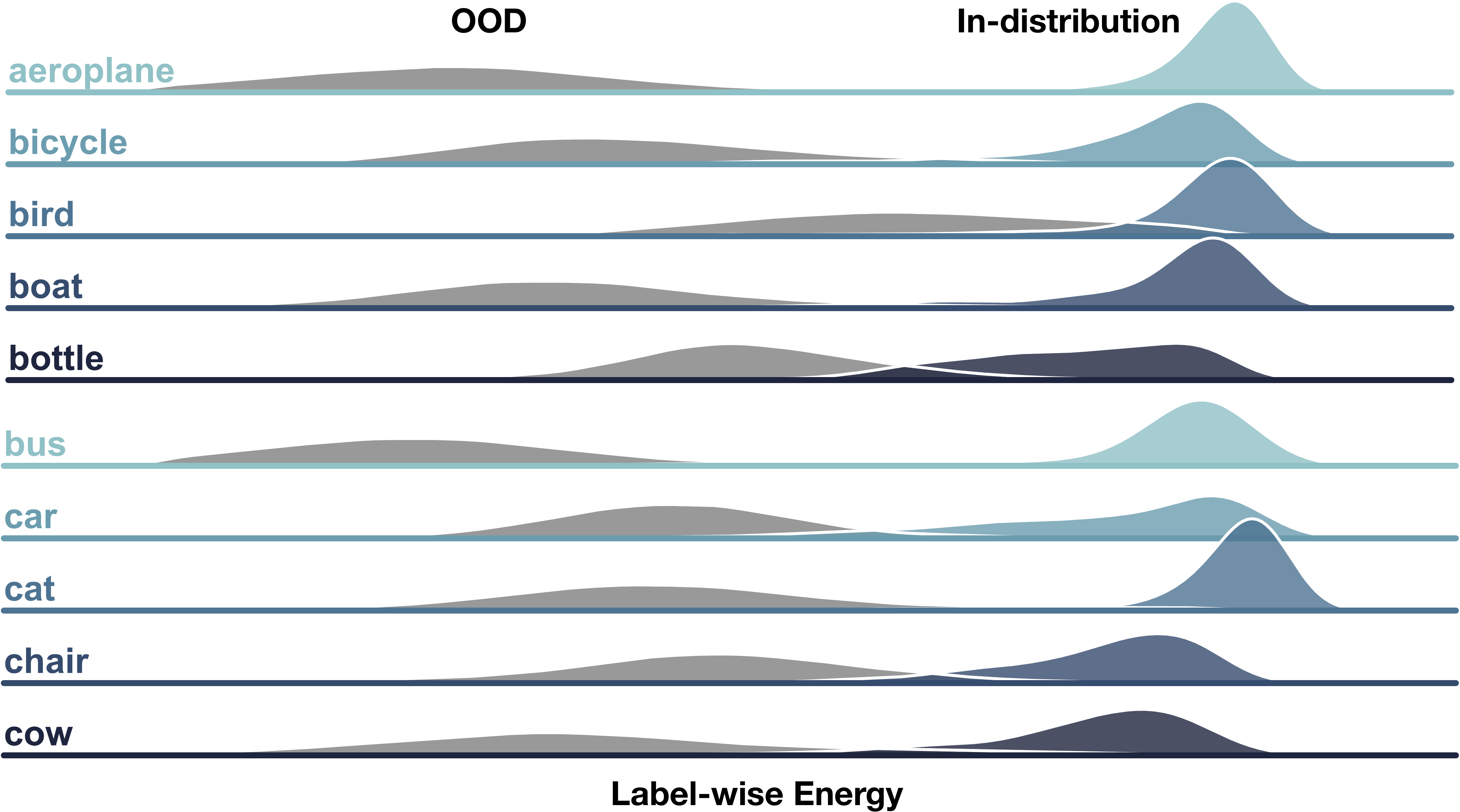}
% \vspace{-0.5cm}
\caption{\footnotesize Label-wise energy scores $-E_{y_i}(\*x)$ distribution. The in-distribution classes (each per row) are a subset from PASCAL-VOC (green). OOD test data is from ImageNet (gray), which is the same for all labels. 
x-axis is in log scale for visibility.
\vspace{-0.5cm}
\label{fig:label_wise}}
\end{center}
\end{figure*}

\textbf{Label-wise Free Energy} We consider a standard pre-trained multi-label neural classifier, with a shared parameter space $\theta$ up to the penultimate feature layer. During inference time, for a given input $\*x$, the (logit) output for the $i$-th class is:
\begin{equation}
    f_{y_i}(\*x) = h(\*x;\theta)\cdot \*w_\text{cls}^i,
\end{equation}
where $\*w_\text{cls}^i$ is the weight vector corresponding to class $i$, and  $h(\*x;\theta)$ is the feature vector in the penultimate layer.  The predictive probability for each binary label $y_i$ is made by a binary logistic classifier:
\begin{equation*}
p(y_i=1 \mid \*x) = \frac{e^{f_{y_i}(\*x)}}{1+ e^{f_{y_i}(\*x)}},
\end{equation*}
where $i \in \{1,2,...,K\}$. 
%For brevity, we use $y_i$ in the probabilistic derivations to indicate the label being positive, \emph{i.e.,} $y_i=1$. 
%For example, we use $p(\*x, y_i)$ to denote the joint probability when $\*x$ is associated with label $y_i=1$.
The logistic classifier output can be viewed as the softmax with two logits---0 and $f_{y_i}(\*x)$, respectively. For each class $y_i$, we define \emph{label-wise free energy} as follows:
\begin{align}\label{eq:class_energy_softmax}
  E_{y_i}(\*x)= -  \text{log} (1+ e^{f_{y_i}(\*x)}),
\end{align}
which can be viewed as a special case of free energy in~\cite{liu2020energy}. For illustration, we show the label-wise energy distribution for a subset of PASCAL-VOC classes in Figure~\ref{fig:label_wise} (green color). Label-wise energy captures the OOD uncertainty for a single label, but unfortunately does not capture uncertainty jointly across labels.

% \textbf{Aggregation Method} We consider three aggregation mechanisms that seek to combine the label-wise energy scores derived above. In particular, \emph{MaxEnergy} indicates the largest label-wise energy score among all labels; \emph{JointEnergy} takes into account the energy scores of all labels.%; and \emph{conditional sum} only considers those labels that are predicted to be positive.  
% \begin{align}
%  \textbf{Max}:~~ E_\text{max}(\*x)& =\max_y E_y(\*x).\\
%   \textbf{Sum}:~~ E_\text{joint}(\*x)& =\sum_{y} E_y(\*x). \label{eq:energy_agg}
%  %  \textbf{Cond. Sum}:~~ E(\*x) & =\sum_{y: f_y(\*x)>=0} E_y(\*x).
% \end{align}
% Now we propose a OOD detector for multi-label classification with $C$ classes as then:
% \begin{align}\label{eq:energy_softmax}
%   E(\*x;f)=- T\cdot \sum_{c}^C \text{log}\sum_{i=1}^2 e^{f^c_i(\*x)/T}.
% \end{align}

\textbf{JointEnergy} We propose a novel scoring function that takes into account the joint uncertainty across labels, and provide mathematical justification from a joint likelihood perspective. In particular, our method is the first to consider the joint estimation of OOD uncertainty across labels:
\begin{align}
% \textbf{Max}:~~ E_\text{max}(\*x)& =\max_i -E_{y_i}(\*x)\\[5pt]
%   \textbf{Sum}:~~ E_\text{joint}(\*x)& =\sum_{i=1}^K -E_{y_i}(\*x)
E_\text{joint}(\*x)& =\sum_{i=1}^K -E_{y_i}(\*x)
  \label{eq:energy_agg}
%  \textbf{ExpSum}:~~ E_\text{expsum}(\*x)& =\sum_{y=1}^K e^{-E_y(\*x)}, 
 %  \textbf{Cond. Sum}:~~ E(\*x) & =\sum_{y: f_y(\*x)>=0} E_y(\*x).
\end{align}
In particular, 
%\emph{MaxEnergy} finds the largest label-wise energy score among all labels; whereas 
\emph{JointEnergy} takes the summation of label-wise energy scores across all labels.
%; \emph{ExpJointEnergy} combines the exponential of energy scores. 
Note that in the above equations, label-wise energy $E_{y_i}(\*x)$ by definition is a {negative} value, and the aggregation methods output a {positive} value by negation. This aligns with the convention that a larger score indicates in-distribution and vice versa. %, particularly \emph{JointEnergy} and \emph{ExpJointEnergy}.

%\SL{Perhaps restructure and make max energy as a separate ablation in the experiments instead?}
% \vspace{2cm}
\textbf{Mathematical Interpretation} 
%We provide mathematical interpretations of different aggregation methods. 
We provide mathematical interpretation for \emph{JointEnergy}.
To interpret \emph{JointEnergy}, we first resort to the energy-based model~\citep{lecun2006tutorial}, where the conditional likelihood $p(\*x \mid y_i=1)$ is given by: 
\begin{align}
\label{eq:conditional-likelihood}
    p(\*x \mid y_i=1) & = \frac{e^{-E_{y_i}(\*x)}}{\int_{\*x | y_i}e^{-E_{y_i}(\*x)}},
\end{align}
and $Z_{y_i}=\int_{\*x \mid {y_i}}e^{-E_{y_i}(\*x)}$ is the normalized density. 
%For notation brevity, we use $p(\*x\mid y_i)$ to indicate the conditional likelihood when $y_i$ is positive, \emph{i.e.}, $y_i=1$. Similarly, we use $p(\*x, y_i)$ to denote the joint probability when $\*x$ is associated with label $y_i=1$. 
% Since $Z_{y_i}$ is the same with respect to all $\*x$ with label $y_i$, the denominator in Equation~\ref{eq:conditional-likelihood} does not affect the overall distributional of label-wise energy scores. 
% By taking the logarithm on both sides of Equation~\ref{eq:conditional-likelihood}, we have:
% \begin{align*}
%     E_{y_i}(\*x) \propto -\log p(\*x \mid y_i),
% \end{align*}
% which suggests that the label-wise energy score is aligned with the conditional likelihood function. 

% Given this, \emph{MaxEnergy} can be interpreted as taking the maximum of conditioned log-likelihood among all labels:  
% \begin{align}
%     E_\text{max}(\*x) \propto \max_i \log p(\*x \mid y_i),
% \end{align}
% which does not take into account the joint information across labels. 

%We provide mathematical interpretation for \emph{JointEnergy}, which is the first method to consider the joint estimation of OOD scores across labels. 
We now show that \emph{JointEnergy} can be interpreted from the joint likelihood perspective:
% Under the assumption that normalized densities $Z_y=\int_{\*x|y}e^{-E_y(\*x)}$ are equal across labels,
\begin{align}
    E_\text{joint}(\*x)&= \sum_{i=1}^K  \log \bigg(p(\*x \mid y_i=1)\cdot Z_{y_i}\bigg) \\[8pt] 
    & = \sum_{i=1}^K  \log p(\*x \mid y_i=1) +\underbrace{\sum_{i=1}^K \log Z_{y_i}}_{Z} \label{eq: sum1}
\end{align}
By applying Bayesian rule for each term $\log p(\*x \mid y_i=1)$ in Equation~\ref{eq: sum1}, we have
\begin{align}

   E_\text{joint}(\*x)&= \log \prod_{i=1}^K  \frac{p(y_i=1 \mid \*x) \cdot p(\*x)}{p(y_i=1)} + Z\\
   %\log \frac{p(\*x, y_1)}{p(y_1)}\cdot \frac{p(\*x, y_2)}{p(y_2)} \cdot \cdot \cdot \frac{p(\*x, y_K)}{p(y_K)} + Z \\[10pt] 
    % &= \log  \prod_{i=1}^K p(\*x, y_i) + \underbrace{(Z - \log \prod_{i=1}^K p(y_i)).}_{\text{\normalfont constant for all  $\*x$}} 
   % &= \log p(\*x) \cdot \frac{p(\*x, y_1)}{p(\*x)}\cdot \frac{p(\*x, y_2)}{p(\*x)} \cdot \cdot \cdot \frac{p(\*x, y_K)}{p(\*x)} \\
  %  & + \log \frac{(p(\*x))^{K-1}}{p(y_1) \cdot p(y_2) \cdot \cdot \cdot p(y_K)} + Z \\[10pt] 
    &= \log \prod_{i=1}^K p(y_i =1\mid \*x)  + K\cdot \log p(\*x) \notag \\
    & \quad + \underbrace{(Z - \log \prod_{i=1}^K p(y_i=1))}_{\text{\normalfont C}} 
    \label{eq:sum2}
\end{align}
Given all labels $y_i$ are conditionally independent\footnote{Note that this is sufficient but not necessary for our results to hold. Our theoretical assumption is made to ease and facilitate the interpretation from a joint likelihood perspective. Importantly, our experiments in Section~\ref{sec:experiments} hold without imposing any condition and demonstrate superior performance. }, we have $\prod_{i=1}^K p(y_i =1\mid \*x) = p(y_1=1,y_2=1,...,y_K=1\mid \*x)$. Therefore, Equation~\ref{eq:sum2} is equivalent to:
\begin{align}
     E_\text{joint}(\*x)&= \log p(y_1=1,y_2=1, \ldots, y_K=1 \mid \*x)
     \notag \\
     & \quad + K\cdot \log p(\*x) + C \\
     \vspace{-0.1in}
     & = \log \frac{p(\*x \mid y_1=1,y_2=1,...,y_K=1)\cdot \prod_{i=1}^K p(y_i=1)}{p(\*x)} \notag \\
     & \quad + K\cdot \log p(\*x)  + C\\
     & =  \underbrace{\log p(\*x \mid y_1=1, y_2=1,...,y_K=1)}_{\text{joint conditional log likelihood}, ~\uparrow~\text{for in-distribution}} 
     \notag \\
     & \quad + \underbrace{(K-1) \cdot \log p(\*x)}_{\text{log data density}, ~\uparrow~\text{for in-distribution}} + ~~~Z \label{eq:interpretation}
    %&= \log p(\*x,y_1,y_2, \ldots, y_K) + (K-1)\log p(\*x) + C
\end{align}
\paragraph{Rationale of Equation~\ref{eq:interpretation}} The equation above suggests that $E_\text{joint}(\*x)$ can be interpreted from the joint conditional log likelihood and log data density perspective. The second term is desirable for OOD detection since it reflects the underlying data density, which is higher for in-distribution data $\*x$. The first term takes into account joint estimation across labels, which is new to our multi-label setting and was not previously considered in multi-class setting~\citep{liu2020energy}. The first term allows even further discriminativity between in- vs. OOD data, since OOD data is expected to have lower joint conditional likelihood (\emph{i.e.}, not associated with any of the labels). In contrast, having multiple dominant labels is indicative of an in-distribution input, which is a characteristic that \emph{JointEnergy} captures.  As a major advantage, our method circumvents the challenge to {directly} estimate the joint likelihood using generative models, which can be computationally intractable to train and optimize on multi-label datasets~\cite{hinz2019generating}.

\subsection{JointEnergy for Multi-Label OOD Detection}
We propose using the JointEnergy function $E_\text{joint}(\*x)$ defined in Section~\ref{sec:method} for OOD detection:
\begin{align}
\label{eq:ood}
G(\*x; \tau) =
\begin{cases}
\text{out} & \quad \text{if } E_\text{joint}(\*x) \leq \tau, \\
\text{in} & \quad \text{if } E_\text{joint}(\*x) > \tau,
\end{cases}
\end{align}
where $\tau$ is the energy threshold, and can be chosen so that a high fraction (\emph{e.g.}, 95\%) of in-distribution data is correctly classified by $G(\*x; \tau)$. The sensitivity analysis on $\tau$ is provided in Figure~\ref{fig:auroc}.\@
%$E(\*x)$ could take on forms of \emph{Max} or \emph{Sum}. 
A data point with higher JointEnergy $E_\text{joint}(\*x)$ is considered as in-distribution, and vice versa (see Fig. \ref{fig:banner}). %We explore and provide the tradeoff of different aggregation methods in Section~\ref{sec:results}.

\begin{table*}[t]
\vspace{-0.5cm}
\centering
\small
\caption[]{\small 
%OOD detection performance comparison using energy-based approaches  vs.\ competitive baselines. We use DenseNet~\citep{huang2017densely} to train on the in-distribution datasets. We use a subset of ImageNet classes as OOD test data, as described in Section~\ref{sec:setup}. All values are percentages. $\uparrow$ indicates larger values are better, and $\downarrow$ indicates smaller values are better. \textbf{Bold} numbers are superior results.
OOD detection performance comparison using JointEnergy  vs.\ competitive baselines. We use DenseNet~\citep{huang2017densely} to train on the in-distribution datasets. We use a subset of ImageNet classes as OOD test data, as described in Section~\ref{sec:setup}. All values are percentages. $\uparrow$ indicates larger values are better, and $\downarrow$ indicates smaller values are better. \textbf{Bold} numbers are superior results. Description of baseline methods, additional evaluation results on different OOD test data, and different architecture (\emph{e.g.}, ResNet~\cite{he2016identity}) can be found in the Appendix.  %\SL{let's move Max energy to be a different small table?}
}

\vspace{0.15in}
\begin{tabular}{lrrr}
\toprule
$\mathcal{D}_{\text{in}}$ & MS-COCO & PASCAL-VOC & NUS-WIDE  \\
 & \multicolumn{3}{c}{{ \textbf{FPR95} / \textbf{AUROC}}  / \textbf{AUPR}} \\
\textbf{OOD Score} & \multicolumn{3}{c}{{ $\downarrow ~~~~~~~ \uparrow~~~~~~~~~~~\uparrow$}} \\

\midrule
{MaxLogit~\citep{hendrycks2019benchmark}}
& 43.53 / 89.11 / 93.74  & 45.06 / 89.22 / 83.14 
& 56.46 / 83.58 / 94.32 \\
% \textbf{MaxProb}
% & 43.53 / 89.11 / 93.74  & 45.06 / 89.22 / 83.14 
% & 56.46 / 83.58 / 94.32 \\
{MSP~\citep{hendrycks2016baseline}}
& 79.90 / 73.70 / 85.37 & 74.05 / 79.32 / 72.54 
& 88.50 / 60.81 / 87.00 \\
{ODIN~\citep{liang2018enhancing}}
& 43.53 / 89.11 / 93.74 & 45.06 / 89.22 / 83.16
& 56.46 / 83.58 / 94.32 \\
{Mahalanobis~\citep{lee2018simple}}
& 46.86 / 88.59 / 93.85 &  41.74 / 88.65 / 81.12
& 62.67 / 84.02 / 95.25 \\
{LOF~\citep{breunig2000lof}}
& 80.44 / 73.95 / 86.01 & 86.34 / 69.21 / 58.93  
& 85.21 / 67.75 / 89.61 \\
{Isolation Forest~\citep{liu2008isolation}} 
& 94.39 / 49.04 / 66.87 & 93.22 / 50.67 / 35.78
& 95.69 / 53.12 / 83.32 \\
\midrule
 %\multirow{2}{0.12\linewidth}{{\textbf{JointEnergy (ours)}}}
% {\textbf{MaxEnergy}}
%  & 43.53 / 89.11 / 93.74 & 45.06 / 89.22 / 83.14
%   & 56.46 / 83.58 / 94.32 \\
% {\textbf{ExpJointEnergy}}
%  & 38.12 / 90.03 / 94.11 & 43.07 / 89.70 / 83.54
%   & 50.44 / 85.84 / 95.02 \\
{\textbf{JointEnergy }} & \textbf{33.48} / \textbf{92.70} / \textbf{96.25} 
 & \textbf{41.01} / \textbf{91.10} / \textbf{86.33} 
 & \textbf{48.98} / \textbf{88.30} / \textbf{96.40}\\
%  {\textbf{SubsetJointEnergy}}
%  & 37.85 / 91.69 / 95.71 & 43.63 / 90.39 / 85.35
%   & 53.00 / 86.35 / 95.68 \\
\bottomrule
\end{tabular}
\vspace{-0.6cm}
\vspace{0.15in}
\label{tab:results-imagenet}
\end{table*}

\vspace{-0.3cm}
\section{Experiments}
\label{sec:experiments}
In this section, we describe our experimental setup (Section~\ref{sec:setup}) and demonstrate the effectiveness of our method on several OOD evaluation tasks (Section~\ref{sec:results}). We also conduct extensive ablation studies and comparative analysis that lead to an improved understanding of different methods. 

\vspace{-0.3cm}
\subsection{Setup}
\label{sec:setup}
\textbf{In-distribution Datasets} We consider three multi-label datasets: MS-COCO ~\citep{lin2014microsoft}, PASCAL-VOC ~\citep{Everingham15}, and NUS-WIDE ~\citep{nus-wide-civr09}. 
% \todo{add citations. Details on the numbers of classes, images}
MS-COCO consists of 82,783 training, 40,504 validation, and 40,775
testing images with 80 common object categories. PASCAL-VOC contains 22,531 images across 20 classes. NUS-WIDE includes 269,648 images across 81 concept labels. Since NUS-WIDE has invalid and untagged images, we follow~\citep{zhu2017learning} and use 119,986 training images and 80,283 test images. %We use the same train/val/test split as \citeauthor{hendrycks2019benchmark}.
% 
%\todo{add image resolution?} 
%\textcolor{blue}{[I mentioned the image resolution in training details part]}\SL{I see!!}

% \SL{somewhere add the pretty plot of label-wise energy distribution}

\textbf{Training Details} 
%\sout{We use a ResNet-101 backbone architecture that is pre-trained on ImageNet-1K. We replace the final layer with 2 fully connected layers, and fine-tune with the logistic sigmoid function for the three multi-label datasets above.} 
We train three multi-label classifiers, one for each dataset above. 
The classifiers have a DenseNet-121 backbone architecture, with a final layer that is replaced by 2 fully connected layers. Each classifier is pre-trained on ImageNet-1K and then fine-tuned with the logistic sigmoid function to its corresponding multi-label dataset.
We use the Adam optimizer~\citep{kingma2014adam} with standard parameters ($\beta_1 = 0.9$, $\beta_2 = 0.999$). The initial learning rate is $10^{-4}$ for the fully connected layers and $10^{-5}$ for convolutional layers. We also augmented the data with random crops and random flips to obtain color images of size $256 \times 256$. After training, the mAP is 87.51\% for PASCAL-VOC, 73.83\% for MS-COCO, and 60.22\% for NUS-WIDE. All experiments are conducted on NVIDIA GeForce RTX 2080Ti.%We provide details on the hyperparameter tuning for baseline detection methods such as ODIN and Mahalanobis in the Appendix. 

\textbf{Out-of-distribution Datasets} To evaluate the models trained on the in-distribution datasets above, we follow the same set up as in~\citep{hendrycks2019benchmark} and use ImageNet~\citep{deng2009imagenet} for its generality. Besides, we evaluate against the Textures dataset~\citep{cimpoi2014describing} as OOD. For ImageNet, we use the same set of 20 classes chosen from ImageNet-22K as in~\citep{hendrycks2019benchmark}. These classes are chosen not to overlap with ImageNet-1k since the multi-label classifiers are pre-trained on ImageNet-1K. Specifically, we use the following classes for evaluating the MS-COCO and PASCAL-VOC pre-trained models: \emph{dolphin, deer, bat, rhino, raccoon, octopus, giant clam, leech, venus flytrap, cherry tree, Japanese cherry blossoms, redwood, sunflower, croissant, stick cinnamon, cotton, rice, sugar cane, bamboo, and turmeric}. Since NUS-WIDE contains high-level concepts like animal, plants and flowers, we use a different set of classes that are distinct from NUS-WIDE: \emph{asterism, battery, cave, cylinder, delta, fabric, filament, fire bell, hornet nest, kazoo, lichen, naval equipment, newspaper, paperclip, pythium, satellite, thumb, x-ray tube, yeast, zither}.

\textbf{Evaluation Metrics} We measure the following metrics that are commonly used for OOD detection: (1)~the false positive rate (FPR95) of OOD examples when the true positive rate of in-distribution examples is at 95\%; (2)~the area under the receiver operating characteristic curve (AUROC); and (3)~the area under the precision-recall curve (AUPR).

\subsection{Results} 
\label{sec:results}

\begin{wrapfigure}{r}{0.45\textwidth}
\vspace{-0.55cm}
\centering
\scriptsize
% \captionsetup{font=scriptsize}
  \includegraphics[width=0.45\textwidth]{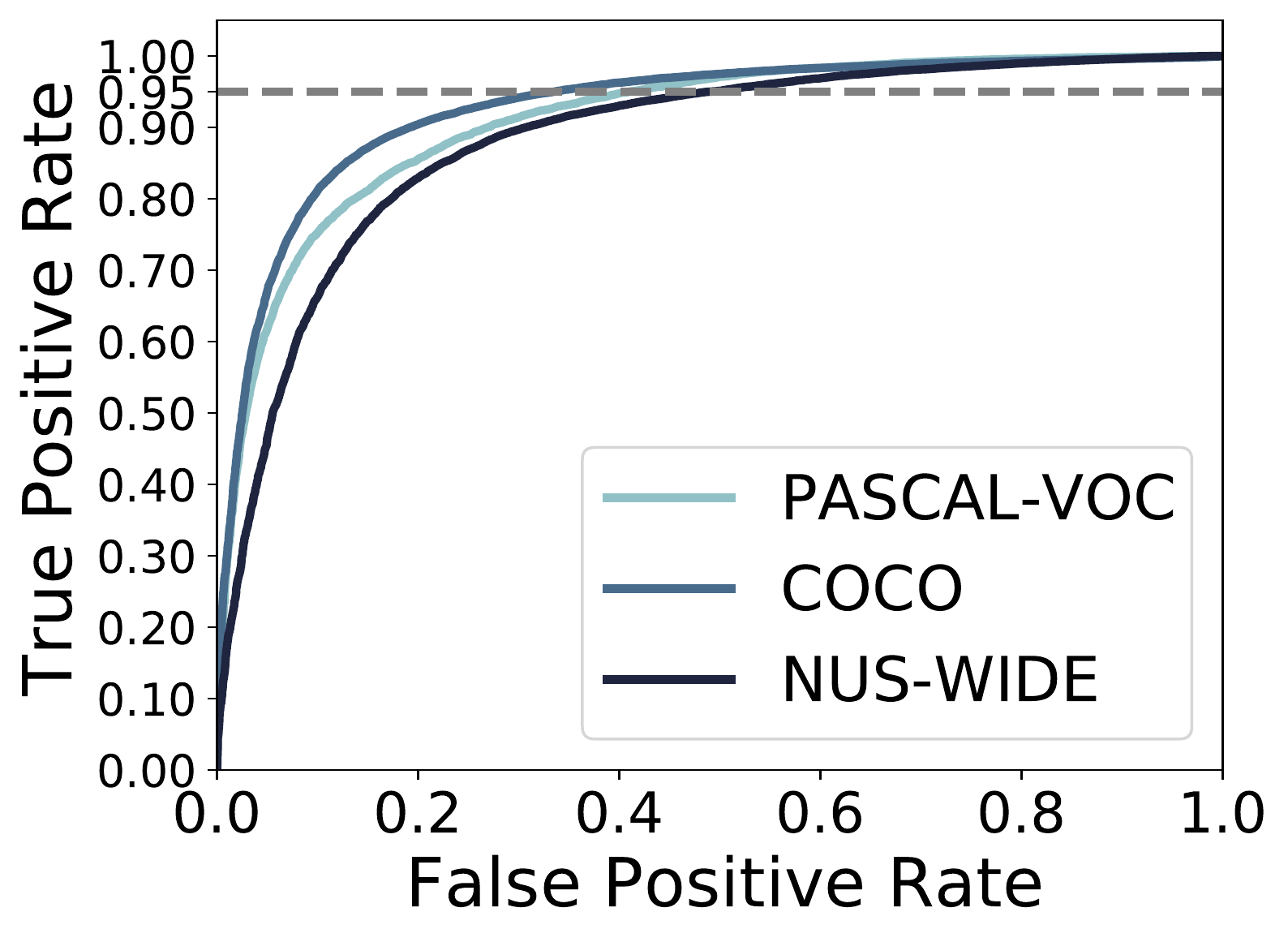}
%   \vspace{-0.7cm}
  \caption{AUROC curves for OOD detector obtained from three in-distribution multi-label classification datasets.} 
%   \SL{update colors to be in similar scheme as main plots, for consistency}}
%   \vspace{-0.5cm}
  \label{fig:auroc}
\end{wrapfigure}

\textbf{How does JointEnergy compare to common OOD detection methods?} In Table~\ref{tab:results-imagenet}, we compare energy-based approaches against competitive OOD detection methods in literature, where \emph{JointEnergy} demonstrates state-of-the-art performance. For fair comparisons, we consider approaches that rely on pre-trained models (without performing retraining or fine-tuning). Following the setup in~\citep{hendrycks2019benchmark},  all the numbers reported are evaluated on ImageNet OOD test data, as described in Section~\ref{sec:setup}. We provide additional evaluation results for the Texture OOD test dataset in the supplementary. 
%Appendix~\ref{sec:appendix}. 
%\SL{missing?}
Most baselines such as MaxLogit~\citep{hendrycks2019benchmark}, Maximum Softmax Probability (MSP)~\citep{hendrycks2016baseline}, ODIN~\citep{liang2018enhancing} and Mahalanobis~\citep{lee2018simple} derive OOD indicator scores based on  the maximum-valued statistics among all labels. Local Outlier Factor (LOF)~\citep{breunig2000lof} uses K-nearest neighbors (KNN) to estimate the local density, where OOD examples are detected from having lower density compared to their neighbors. Isolation forest~\citep{liu2008isolation} is a tree-based approach, which detects anomaly based on the path length from the root node to the terminating node. 
% \todo{Please complete this..}

% \begin{wrapfigure}{hr}{0.4\columnwidth}
% \vspace{-0.4cm}
% \scriptsize
% % \captionsetup{font=scriptsize}
%   \includegraphics[width=0.23\textwidth]{tex/fig/roc.pdf}
% %   \vspace{-0.7cm}
%   \caption{AUROC curves for OOD detector obtained from three in-distribution multi-label classification datasets.}
% %   \vspace{-0.5cm}
%   \label{fig:auroc}
% \end{wrapfigure}

\begin{table*}[t]
\vspace{-0.5cm}
\centering
\caption[]{\small Ablation study on the effect of summation for prior approaches. We use DenseNet~\citep{huang2017densely} to train on the in-distribution datasets. We use ImageNet as OOD test data as described in Section~\ref{sec:setup}. Note that \emph{Sum} does not apply to tree-based or KNN-based approaches (\emph{\emph{\emph{\emph{e.g.}}}}, LOF and Isolation Forest).  % \SL{add citations here to other methods?} \textcolor{blue}{[Since we use different aggregation methods, not sure whether is reasonable to add citations here.]}
}
\small
\vspace{0.15in}
\begin{tabular}{llrrr}
\toprule
&  $\mathcal{D}_{\text{in}}$ & MS-COCO & PASCAL & NUS-WIDE \\
&  & \multicolumn{3}{c}{{ \textbf{FPR95} / \textbf{AUROC} / \textbf{AUPR}}} \\
\textbf{OOD Score} & \textbf{Aggregation} & \multicolumn{3}{c}{{ $\downarrow$ ~~~~~~~ $\uparrow$~~~~~~~~~~$\uparrow$}}\\
\midrule
{Logit} & Sum&
95.46 / 61.81 / 80.39 & 87.18 / 72.68 / 61.24 
  & 96.53 / 51.75 / 82.55 \\
{Prob} & Sum& 
{45.04 / 89.32 / 94.40} & \textbf{38.57} / 86.53 / 79.10
& 50.84 / 83.82 / 95.15 \\
% \textbf{Logit} & Cond. Sum\\
{ODIN} & Sum &
{56.56 / 84.62 / 92.24} & 50.35 / 79.45 / 70.19
& {56.26 / 81.04 / 94.34}\\
{Mahalanobis} & Sum 
& 53.43 / 87.52 / 93.35 
& 44.43 / 87.76 / 79.86
& 69.05 / 80.46 / 94.09\\
{LOF} & Sum & N/A & N/A & N/A \\
{Isolation Forest} & Sum & N/A & N/A & N/A \\
\midrule
\textbf{JointEnergy (ours)} & Sum & \textbf{33.48} / \textbf{92.70} / \textbf{96.25} & 41.01 / \textbf{91.10} / \textbf{86.33} & \textbf{48.98} / \textbf{88.30} / \textbf{96.40}\\
% \textbf{ODIN} & Cond. Sum \\
\bottomrule
\end{tabular}
\vspace{0.15in}
\vspace{-0.6cm}
\label{tab:ablation-imagenet}
\end{table*}

Among different approaches, \emph{JointEnergy} outperforms the best-performing baseline across all three multi-label classifiers considered. In particular, on a network trained with the MS-COCO dataset, \emph{JointEnergy} reduces FPR95 by \textbf{10.05}\%, compared to MaxLogit. We provide the AUROC curves for our method JointEnergy in Figure~\ref{fig:auroc}, for all three in-distribution datasets considered. The y-axis is the true positive rate (TPR), whereas the x-axis is the FPR. The curves indicate how the OOD detection performance changes as we vary the threshold $\tau$ in Equation~\ref{eq:ood}. {We additionally evaluate on a different architecture, ResNet~\citep{he2016identity}, for which we observe consistent improvement and provide details in the supplementary.}

We also note here that existing approaches (such as Mahalanobis distance~\citep{lee2018simple}) requires training a separate classifier for OOD detection. In contrast, JointEnergy  is hyperparameter-free and easy to use in practice. In particular, the Mahalanobis approach is based on the assumption that feature representation forms class-conditional Gaussian distributions, and hence may not be well suited for the multi-label setting (which requires joint distribution to be learned).

% \begin{figure}[h]
%      %\centering
%         \subfigure[MS-COCO]{
%         \begin{minipage}[t]{0.3\linewidth}
%         \centering
%         \includegraphics[width=4.8cm]{tex/fig/ROC/coco.pdf}
%         %\caption{fig2}
%         \end{minipage}%
%         }%
%     \hfill
%         \subfigure[PASCAL-VOC]{
%         \begin{minipage}[t]{0.3\linewidth}
%         \centering
%         \includegraphics[width=4.8cm]{tex/fig/ROC/pascal.pdf}
%         %\caption{fig1}
%         \end{minipage}%
%         }%
%     \hfill
%         \subfigure[NUS-WIDE]{
%         \begin{minipage}[t]{0.3\linewidth}
%         \centering
%         \includegraphics[width=4.8cm]{tex/fig/ROC/nus-wide.pdf}
%         %\caption{fig2}
%         \end{minipage}%
%         }% 
%         \caption{AUROC curves for OOD detector obtained from three in-distribution datasets.}
%     \label{fig:auroc}
% \end{figure}

% \begin{figure}
%     \centering
%     \includegraphics[width=8cm]{tex/fig/ROC/roc.pdf}
%     \label{fig:my_label}
% \end{figure}

\textbf{How do different aggregation methods affect OOD detection performance?}
In Table~\ref{tab:max-sum}, we also perform a comparative analysis of the effect of different aggregation functions that combine label-wise energy scores. As an alternative, we consider \begin{align}
E_\text{max}(\*x)& =\max_i -E_{y_i}(\*x),
%   \textbf{Sum}:~~ E_\text{sum}(\*x)& =\sum_{i=1}^K -E_{y_i}(\*x)
  \label{eq:energy_max}
%  \textbf{ExpSum}:~~ E_\text{expsum}(\*x)& =\sum_{y=1}^K e^{-E_y(\*x)}, 
 %  \textbf{Cond. Sum}:~~ E(\*x) & =\sum_{y: f_y(\*x)>=0} E_y(\*x).
\end{align}
which finds the largest label-wise energy score among all labels.
We observe that \emph{MaxEnergy} does not outperform \emph{JointEnergy}, which utilizes information jointly from all the labels. The performance of \emph{MaxEnergy} is on par with MaxLogit since \emph{MaxEnergy}, given by $\max_i \log(1+e^{f_{y_i}(\*x)})$, is approximately close to the MaxLogit when $f_{y_i}(\*x)$ is large. The results underline the importance of taking into account information from multiple labels, not just the maximum-valued label. This is because, in multi-label classification, the model may assign high probabilities to several classes. Theoretically, \emph{JointEnergy} is also more meaningful, and can be interpreted from a joint likelihood perspective as shown in Section~\ref{sec:method}.%If we simply take the maximum value as the score, which is the same as processing an image containing only one object with the highest probability, we would miss out on other information. 

\begin{table}[b]
\vspace{-0.5cm}
\centering
\small
\caption[]{\small Ablation study on the effect of aggregation methods: max vs summation. Values are AUROC.
}
\vspace{0.15in}
\begin{tabular}{lrr}
\toprule
$\mathcal{D}_{\text{in}}$  & MaxEnergy & JointEnergy  \\
%   \multicolumn{3}{c}{{ \textbf{FPR95} / \textbf{AUROC}}  / \textbf{AUPR}} \\
% $\mathcal{D}_{\text{in}}$ \qquad \multicolumn{3}{c}{{ $\downarrow ~~~~~~~ \uparrow~~~~~~~~~~~\uparrow$}} \\
\midrule
\textbf{MS-COCO}
& 89.11   &  \textbf{92.70}  \\
\textbf{PASCAL-VOC}
&  89.22 & \textbf{91.10}  \\
\textbf{NUS-WIDE}
& 83.58  & \textbf{88.30}\\
\bottomrule
\end{tabular}
% \vspace{-0.2cm}
\vspace{0.15in}
\label{tab:max-sum}
\end{table}

% \begin{table*}[t]
% \vspace{-0.5cm}
% \centering
% \small
% \caption[]{\small Ablation study on the effect of aggregation methods: max vs summation.
% }
% \vspace{0.15in}
% \begin{tabular}{lrr}
% \toprule
% \textbf{OOD Score} & MaxEnergy & JointEnergy  \\
%   \multicolumn{3}{c}{{ \textbf{FPR95} / \textbf{AUROC}}  / \textbf{AUPR}} \\
% $\mathcal{D}_{\text{in}}$ \qquad \multicolumn{3}{c}{{ $\downarrow ~~~~~~~ \uparrow~~~~~~~~~~~\uparrow$}} \\
% \midrule
% \textbf{MS-COCO}
% & 43.53 / 89.11 / 93.74  & \textbf{33.48} / \textbf{92.70} / \textbf{96.25}  \\
% \textbf{PASCAL-VOC}
% & 45.06 / 89.22 / 83.14 & \textbf{41.01} / \textbf{91.10} / \textbf{86.33} \\
% \textbf{NUS-WIDE}
% & 56.46 / 83.58 / 94.32 & \textbf{48.98} / \textbf{88.30} / \textbf{96.40} \\
% \bottomrule
% \end{tabular}
% % \vspace{-0.2cm}
% \vspace{0.15in}
% \label{tab:max-sum}
% \end{table*}

% %Interestingly, we also observe that \emph{ExpJointEnergy} performs favorably compared to \emph{JointEnergy}, though slightly suboptimal. 
% As seen in Section~\ref{sec:agg_energy}, %Since the normalized densities $Z_y$ can be intractable  to compute in practice, our aggregation method \emph{JointEnergy} is based on the simplified scenario where all normalized densities are equal across labels. We note here that this might not strictly hold true on the real-world datasets evaluated. Despite this, 
% We show that \emph{JointEnergy} is effective on all three multi-label classifiers, which supports its practical utility.  

\textbf{What is the effect of applying the aggregation method to prior methods?} As an extension, we explore the effectiveness of applying the aggregation method to previous scoring functions. The results are summarized in Table~\ref{tab:ablation-imagenet}. We calculate scores based on the logit $f_{y_i}(\*x)$, sigmoid of the logit $\frac{1}{1+e^{-f_{y_i}(\*x)}}$, ODIN score, and %\textcolor{red}{weitang: from google: 'As well as is a conjunction (a connecting word or phrase) that means 'in addition to. ' It does not mean the same thing as 'and,' but implies that one of the items in a list deserves emphasis.'  I will just use and } 
Mahalanobis distance score $M_{y_i}(\*x)$ for each label independently. We then perform summation across the label-wise scores as the overall OOD score. This ablation essentially replaces the \emph{Max} aggregation with \emph{Sum}, which helps understand the extent to which previous approaches are amenable in the multi-label setting. Note that the summation aggregation method does not apply to tree-based or KNN-based approaches such as LOF and Isolation Forest.

% \begin{wrapfigure}{0.5\linewidth}
%     \centering
%     \vspace{-30mm}
%     \includegraphics[width=0.45\linewidth]{tex/fig/energy-ood-pascal-crop.pdf}
%     \caption{\footnotesize Label-wise energy scores $-E_y(\*x)$ for in-distribution example from PASCAL-VOC (top), and OOD input from ImageNet (bottom). The OOD input is misclassified using MaxLogit score since the dominant output has high an activation, and makes it less distinguishable from in-distribution data's MaxLogit score. In contrast, \emph{JointEnergy} correctly classify both images since it results in larger differences in scores between in-dist. and OOD inputs.  }
%      \vspace{-5mm}
%     \label{fig:example}
% \end{wrapfigure}

\begin{figure*}[t]
\begin{center}  
\includegraphics[width=140mm]{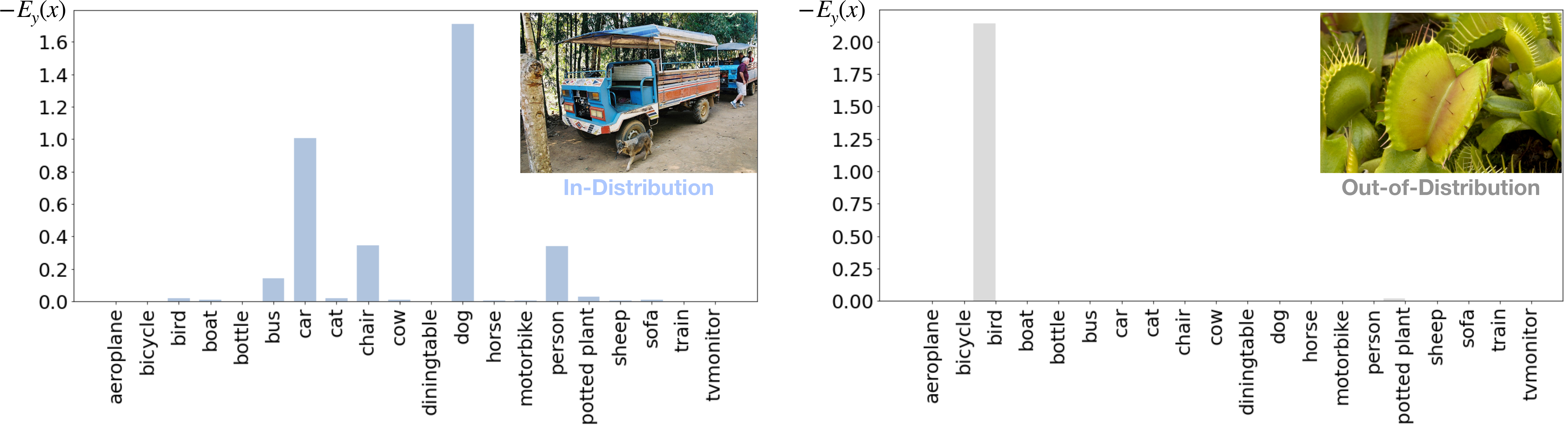}
% \vspace{-0.5cm}
\caption{\footnotesize Label-wise energy scores $-E_{y_i}(\*x)$ for in-distribution example from PASCAL-VOC (left), and OOD input from ImageNet (right). The OOD input is misclassified using MaxLogit score since the dominant output has a high activation, making it indistinguishable from an in-distribution data's MaxLogit score. In contrast, \emph{JointEnergy} correctly classifies both images since it results in larger differences in scores between in-distribution and OOD inputs.  
\vspace{-0.6cm}
\label{fig:example}}
\end{center}
\end{figure*}
We found that applying summation over individual logit/MSP/ODIN/Mahalanobis scores from each label does not enhance but sometimes worsen the performance. For example, simply summing over the logits across the labels leads to severe degradation in performance since the outputs are mixed with positive and negative numbers. On MS-COCO, the FPR degrades from 43.53\% using MaxLogit to 95.46\% (using SumLogit). In contrast, \emph{JointEnergy} does not suffer from this issue. 
%More importantly, the label-wise scores derived from energy is theoretically more meaningful than logit/MSP/ODIN/Mahalanobis, since it is provably aligned with the probability density of the training data corresponding to the label. 
This underlines the importance of choosing proper label-wise scoring function to be compatible with the aggregation method. 
%\todo{By default those method should be viewed as taking max as aggregation method. Highlight the fact that logit/MSP/ODIN fails under sum. Mahalanobis is not applicable because it's in distance space. These method fail because...}

\textbf{JointEnergy vs. SumProb} We highlight the advantage of JointEnergy over SumProb both empirically and theoretically. As seen in Table~\ref{tab:ablation-imagenet}, the performance difference between JointEnergy and SumProb is substantial. In particular, on MS-COCO, our method outperforms SumProb by 11.56\% (FPR95). For threshold independent metric AUROC, JointEnergy consistently outperforms SumProb by 3.38\% (MS-COCO), 4.57\% (PASCAL), and 4.48\% (NUS-WIDE). 

% JointEnergy is a mathematically meaningful measurement and can be interpreted from a joint likelihood perspective (see Section~\ref{sec:agg_energy}), whereas SumProb does not. In fact, one can show that the probability score for each individual label is not aligned with the conditioned data density function. To see this, we can derive the probability for each binary logistic classifier as:
% \begin{align*}
% 	\log p(y_i \mid \*x) &=  \log \frac{e^{f_{y_i}(\*x)}}{1+ e^{f_{y_i}(\*x)}} = f_{y_i}(\*x) + E_{y_i}(\*x).
% \end{align*}
% In fact, the first term $f_{y_i}(\*x)$ is larger for in-distribution data with label $y_i$, whereas the second term is smaller for in-distribution data. This leads to a \emph{biased scoring distribution} that is no longer proportional to the label-conditional log-likelihood $\log p(\*x\mid y_i)$, which corroborates~\cite{liu2020energy}. %:
% % \begin{align*}
% % 	\log p(y_i \mid \*x) \not \propto \text{class-conditional likelihood of data with label}~y_i
% % \end{align*}
% SumProb, as a result, inherits this weakness theoretically and performs worse than JointEnergy. 

\textbf{Qualitative Case Study} Lastly, to provide further insights on our method, we qualitatively examine examples from the multi-label classification dataset PASCAL-VOC (in-dist.) and OOD input from the ImageNet that are correctly classified by \emph{JointEnergy} but not MaxLogit. In Figure~\ref{fig:example} (left), we see an in-distribution example is labeled as \texttt{dog}, \texttt{car}, \texttt{chair} and \texttt{person}, with \emph{MaxLogit} score 1.63 and \emph{JointEnergy} score 3.23. We also show an OOD input (Figure~\ref{fig:example}, right) with a single dominant activation on the \texttt{bird} class, with MaxLogit score 2.14 and \emph{JointEnergy} score 2.19.  In this example, taking the sum appropriately results in a higher score for the in-distribution image than the OOD image. Contrarily, MaxLogit score for the in-distribution image is lower than that of the OOD image, which results in ineffective detection. 

\section{Related Work}
\label{sec:related}

\paragraph{Multi-label Classification}
%\textcolor{red}{Wrote about some examples of multi-label applications, not sure if I should write more about methods in multi-label classification?}
The task of identifying multiple classes within an input example is of significant interest in many applications~\citep{tsoumakas2007multi} where deep neural networks are commonly used as the classifiers. Natural images usually contain several objects and may have many associated tags~\citep{wang2016cnn}. 
%\citep{gong2013deep} \textcolor{red}{weitang:I think citations like [number] cannot be used as subjects as John mentioned before. 
Chen \emph{et al.}~\cite{gong2013deep} used convolutional neural networks (CNN) to annotate images with 3 or 5 tags on the NUS-WIDE dataset. \citep{chen2019integration} used CNNs to tag images of road scenes from 52 possible labels. %In the medical domain, multi-label classification is used for detecting several abnormalities that may exist in one image. 
In the medical domain, Wang \emph{et al.}~\cite{wang2017chestx} presented a chest X-ray dataset in which one image may contain multiple abnormalities. Multi-label classification is also prominent in natural language processing~\citep{nam2014large}.\@ %The dataset originally has 8 abnormalities, which was later expanded to 14. 
%They evaluate various deep learning architectures for detecting the presence of any abnormalities in an image. 
Our proposed method is therefore relevant to a wide range of applications in the real world.% such as for identifying unexpected objects in a road scene \citep{blum2019fishyscapes}. %, \textcolor{red}{or detecting abnormalities in a medical image (should we mention this?)}.

\textbf{Out-of-distribution Uncertainty Estimation}  
% The softmax confidence score has become a common baseline for OOD detection~\citep{hendrycks2016baseline}.
% A theoretical investigation~\citep{hein2019relu}  shows that neural networks with ReLU activation can produce arbitrarily high softmax confidence for OOD inputs. 
Detecting and rejecting unknowns has a long history in machine learning; see \cite{yang2021oodsurvey} for a comprehensive survey of the main ideas. We highlight a few representative works in the context of deep learning. The phenomenon of neural networks' overconfidence to out-of-distribution data is revealed by Nguyen \emph{et al.}~\cite{nguyen2015deep}.
Previous works attempt to improve the OOD uncertainty estimation by proposing the ODIN score~\citep{hsu2020generalized, liang2018enhancing}, Mahalanobis
distance-based confidence score~\citep{lee2018simple}, and gradient-based GradNorm score~\citep{huang2021importance}.
% Deep ensembles~\cite{lakshminarayanan2017simple} effectively improve OOD uncertainty estimation at the cost of extra computation. ODIN~\cite{liang2018enhancing} can improve the performance of OOD detectors using calibrated softmax scores with temperature scaling~\cite{guo2017calibration}. In addition, they apply pre-processing input data with adversarial perturbations~\cite{goodfellow2014explaining} to make the maximum softmax probability more distinguishable between in- and out-of- distribution examples.
%\citep{devries2018learning} propose to learn the confidence score by using an auxiliary branch
%to derive the OOD score. 
Recent work by Liu \emph{et al.}~\citep{liu2020energy} proposed using an energy score for OOD detection, which demonstrated advantages over the softmax confidence score both empirically and theoretically. Huang and Li~\cite{huang2021mos} proposed a group-based OOD detection method that scales effectively to large-scale dataset ImageNet. 
%Lin \emph{et al.}~\cite{lin2021mood} proposed the first dynamic OOD inference framework that improved
%the computational efficiency of OOD detection. 
However, previous methods primarily focused on  multi-class classification networks. In contrast, we propose a hyperparameter-free measurement that allows effective OOD detection in the underexplored \emph{multi-label} setting, where the information from  various labels is combined in a theoretically interpretable manner.

\textbf{Generative-based Out-of-distribution Detection} Generative models \citep{dinh2016density,huang2017stacked, kingma2013auto,rezende2014stochastic,tabak2013family,van2016conditional} can be alternative approaches for detecting OOD examples, as they directly estimate the in-distribution density and can declare a test sample to be out-of-distribution if it lies in the low-density regions. However, as shown by Nalisnick \emph{et al.}~\cite{nalisnick2018deep}, deep generative models can assign a high likelihood to out-of-distribution data.
% , which urges caution against using deep generative models to detect out-of-distribution inputs.
Deep generative models can be more effective for out-of-distribution detection using likelihood ratio test~\citep{ren2019likelihood, serra2019input} and likelihood regret~\citep{xiao2020likelihood}.
Though our work is based on discriminative classification models, we show that label-wise energy scores can be theoretically interpreted from a data density perspective. More importantly, generative based models~\citep{hinz2019generating} can be prohibitively challenging to train and optimize, especially on large and complex multi-label datasets that we considered (\emph{e.g.}, MS-COCO, NUS-WIDE etc.). In contrast, our method relies on a discriminative multi-label classifier, which can be easily optimized using standard SGD. %Our method  therefore inherits the merits of  generative-based approaches, while circumventing the difficult optimization process in training generative models, especially on large, complex input data.
%Note that we mainly considered discriminative-based approaches, which can be more competitive due to the availability of label information (and, in some cases, auxiliary outlier data~\cite{hein2019relu, hendrycks2018deep,meinke2019towards, mohseni2020self}). 

\textbf{Energy-based Learning} Energy-based machine learning models date back to Boltzmann machines~\citep{ackley1985learning, salakhutdinov2010efficient}. %, networks of units with energy defined for the overall network.
Energy-based learning~\citep{lecun2006tutorial,ranzato2007efficient, ranzato2007unified} provides a unified framework for many probabilistic and non-probabilistic approaches to learning. Recent work~\citep{zhao2019energy} also demonstrated using energy functions to train GANs~\citep{goodfellow2014generative}, where the discriminator uses energy values to differentiate between real and generated images.
% Recent work~\cite{grathwohl2019your} incorporated energy-based model to estimate the joint distribution from a generative modeling perspective. The downstream OOD detection used standard probabilistic scores. Unlike ours, we show that non-probabilistic energy scores can be directly used as a scoring function for OOD uncertainty estimation. Our optimization objective fits more naturally within energy-based model than generative models, and facilitates the use of modern discriminative neural architectures. 
Xie \emph{et al.}~\cite{xie2016theory} showed that a discriminative classifier can be interpreted from an energy-based perspective. 
%Subsequent  works~\citep{xie2017synthesizing, xie2019learning, xie2018learning, xie2018cooperative} explored video generation, 3D shape pattern generation, and text generation~\citep{deng2019residual} through EBMs. 
Energy-based methods are also used in structure prediction \citep{belanger2016structured, tu2018learning}. 
% Grathwohl \emph{et al.}~\cite{grathwohl2019your} showed that a discriminative classifier can be interpreted from an energy-based perspective. The proposed JEM optimization objective estimates the joint distribution $p(\*x,y)$ from a generative perspective, which requires estimating the normalized densities and can be intractable and unstable to compute. 
Liu \emph{et al.}~\cite{liu2020energy} first proposed using energy score for OOD uncertainty estimation, which demonstrated superior performance for multi-class classification networks. In contrast, our work focuses on a  multi-label setting, where we contribute both empirical and theoretical insights and demonstrate the effectiveness of utilizing information jointly from across all labels.

\section{Conclusion and Outlook}
\label{sec:conclusion}

In this work, we propose energy scores for OOD uncertainty estimation in the multi-label classification setting. We show that aggregating energies over all labels into \emph{JointEnergy} results in better separation between in-distribution and OOD inputs compared to using information from only one label's information. Additionally, we justify the mathematical interpretation of \emph{JointEnergy} from a joint likelihood perspective. \emph{JointEnergy} obtains better OOD detection performance compared to competitive baseline methods, establishing new state-of-the-art on this task. Applications of multi-label classification can benefit from our methods, and we anticipate further research in OOD detection to extend this work to tasks beyond image recognition. We hope our work will increase the attention toward a broader view of OOD uncertainty estimation for multi-label classification.

\section{Societal Impact}
\label{sec:broad}
% \vspace{-0.3cm}
Our project aims to improve the dependability and trustworthiness of modern machine learning models for multi-label classification.
This stands to benefit a wide range of fields and societal activities. We believe out-of-distribution uncertainty estimation is an increasingly critical component of systems that range from consumer and business applications (e.g., digital content understanding) to transportation (e.g., driver assistance systems and autonomous vehicles), and to health care (e.g., unseen disease identification). Many of these applications require multi-label classification models in operation.\@
Through this work and by releasing our code, we hope to provide machine learning researchers a new methodological perspective and offer machine learning practitioners an easy-to-use tool that renders safety against OOD data in the real world. While we do not anticipate any negative consequences to our work, we hope to continue to build on our framework in future work. 

\section*{Acknowledgement}
 Research is supported by the Office of the Vice Chancellor for Research and Graduate Education (OVCRGE) with funding from the Wisconsin Alumni Research Foundation (WARF).
\newpage

\bibliography{icml2021_conference}
\bibliographystyle{plain}

\clearpage
% \onecolumn
% \icmltitle{Supplemental Materials: Towards Uncertainty Estimation for Multi-label Classification in Open World
% }
\appendix
\section*{Appendix}
\label{sec:appendix}

\section{Evaluation on different architecture}

We provide additional evaluation results for ResNet~\citep{he2016identity}. 
The classifiers have a ResNet-101 backbone architecture, but with a final layer that is replaced by 2 fully connected layers. Each classifier is pre-trained on ImageNet-1K and then fine-tuned with the logistic sigmoid function to its corresponding multi-label dataset. We use the same training settings as in the main paper. After training, the mAP is 87.73\% for PASCAL-VOC, 72.77\% for MS-COCO, and 61.47\% for NUS-WIDE.

In Table~\ref{tab:resnet-imagenet}, we show the performance comparison of various OOD detection approaches, evaluated on ImageNet as the OOD test set. The ablation of applying summation over baseline methods is provided in Table~\ref{tab:resnet-ablation-imagenet}.

\begin{table}[h]
\vspace{-0.2cm}
\centering
\small
\caption[]{\small OOD detection performance comparison using energy-based approaches  vs.\ competitive baselines. We use ResNet~\citep{he2016identity} to train on the in-distribution datasets. We use a subset of ImageNet classes as OOD test data, as described in Section~\ref{sec:setup}. All values are percentages. $\uparrow$ indicates larger values are better, and $\downarrow$ indicates smaller values are better. \textbf{Bold} numbers are superior results. 
}
\begin{tabular}{lrrr}
\toprule
 $\mathcal{D}_{\text{in}}$ & MS-COCO & PASCAL-VOC & NUS-WIDE  \\
& \multicolumn{3}{c}{{ \textbf{FPR95} / \textbf{AUROC}}  / \textbf{AUPR}} \\
\textbf{OOD Score}  & \multicolumn{3}{c}{{ $\downarrow ~~~~~~~ \uparrow~~~~~~~~~~~\uparrow$}} \\

\midrule
{MaxLogit}~\cite{hendrycks2019benchmark}
&34.54 / 90.93 / 94.30 & 36.32 / 91.04 / 82.68 
& 58.05 / 83.07 / 94.21\\
% {MaxProb}~\cite{hendrycks2016baseline}
% &34.54 / 90.93 / 94.30 & 36.32 / 91.04 / 82.68 
% & 58.05 / 83.07 / 94.21\\
{MSP}~\cite{hendrycks2016baseline}& 77.92 / 72.43 / 84.34 & 69.85 / 78.24 / 67.93 & 88.75 / 59.19 / 86.40\\
{ODIN}~\cite{liang2018enhancing}  & 34.58 / 90.26 / 93.69   &36.32 / 91.04 / 82.68 & 58.05 / 83.07 / 94.21\\
{Mahalanobis}~\cite{lee2018simple}
& 94.04 / 49.49 / 70.71  & 78.02 / 70.93 / 59.84
& 61.33 / 83.75 / 95.15\\
{LOF}~\cite{breunig2000lof}
&74.30 / 74.87 / 85.82 & 76.71 / 67.54 / 55.35 
& 85.42 / 69.37 / 90.36\\
{Isolation Forest}~\cite{liu2008isolation} 
& 99.06 / 37.59 / 63.43 &98.64 / 41.94 / 33.50 
& 96.59 / 50.75 / 82.91\\
\midrule
%  \multirow{2}{0.12\linewidth}{{\textbf{Energy (ours)}}}
%   &  Max & 34.54 / 90.93 / 94.30 & 36.32 / 91.04 / 82.68 & 58.05 / 83.07 / 94.21\\
%  & ExpSum & 33.64 / 91.07 / 94.38 & 34.79 / 91.36 / 83.10 & 51.87 / 85.58 / 95.02 \\
\textbf{ JointEnergy} (ours)  & \textbf{31.51 / 92.68 / 96.15} & \textbf{31.96 / 92.32 / 86.87} & \textbf{50.25 / 88.12 / 96.34}\\
\bottomrule
\end{tabular}
% \vspace{-0.2cm}
% \caption[]{\small OOD detection performance comparison using energy-based approaches  vs.\ competitive baselines. We use ResNet~\citep{he2016identity} to train on the in-distribution datasets. We use a subset of ImageNet classes as OOD test data, as described in Section~\ref{sec:setup}. All values are percentages. $\uparrow$ indicates larger values are better, and $\downarrow$ indicates smaller values are better. \textbf{Bold} numbers are superior results. 
% }
\label{tab:resnet-imagenet}
\end{table}

\begin{table}[h]
\centering
\small
\caption[]{\small Ablation study on the effect of aggregation methods for prior approaches. We use ResNet~\citep{he2016identity} to train on the in-distribution datasets. We use ImageNet as OOD test data as described in Section~\ref{sec:setup}. Note that \emph{Sum} is not applicable to tree-based or KNN-based approaches (\emph{e.g.}, LOF and Isolation Forest).   
}
\begin{tabular}{llrrr}
\toprule
&  $\mathcal{D}_{\text{in}}$ & MS-COCO & PASCAL & NUS-WIDE \\
&  & \multicolumn{3}{c}{{ \textbf{FPR95} / \textbf{AUROC} / \textbf{AUPR}}} \\
\textbf{OOD Score} & \textbf{Aggregation} & \multicolumn{3}{c}{{ $\downarrow$ ~~~~~~~ $\uparrow$~~~~~~~~~~$\uparrow$}}\\
\midrule
\textbf{Logit} & Sum&
95.63 / 53.52 / 73.25 & 96.36 / 49.44 / 43.07 & 96.49 / 49.83 / 81.78\\
\textbf{Prob} & Sum& 
{43.69 / 87.21 / 93.14} & {35.97 / 84.68 / 76.61} & 55.86 / 82.97 / 94.92\\
% \textbf{Logit} & Cond. Sum\\
\textbf{ODIN} & Sum &
{43.69 / 87.21 / 93.14} & 53.77 / 74.50 / 67.15 & {55.24 / 81.84 / 94.59}\\
\textbf{Mahalanobis} & Sum & 94.47 / 46.82 / 67.06 
& 78.56 / 70.84 / 59.34  
& 62.79 / 83.19 / 94.96\\
\textbf{LOF} & Sum & N/A & N/A & N/A \\
\textbf{Isolation Forest} & Sum & N/A & N/A & N/A \\
\midrule
\textbf{Energy}  & Sum & \textbf{31.51 / 92.68 / 96.15} & \textbf{31.96 / 92.32 / 86.87} & \textbf{50.25 / 88.12 / 96.34}\\
% \textbf{ODIN} & Cond. Sum \\
\bottomrule
\end{tabular}
% \vspace{-0.2cm}
% \caption[]{\small Ablation study on the effect of aggregation methods for prior approaches. We use ResNet~\citep{he2016identity} to train on the in-distribution datasets. We use ImageNet as OOD test data as described in Section~\ref{sec:setup}. Note that \emph{Sum} is not applicable to tree-based or KNN-based approaches (e.g., LOF and Isolation Forest).   
% }
\label{tab:resnet-ablation-imagenet}
\end{table}

\section{Evaluation on different OOD test data}
In addition to ImageNet, we also evaluate on a different OOD test dataset, Textures~\citep{cimpoi14describing}. The results are reported in Table~\ref{tab:results-texture} and Table~\ref{tab:ablation-texture}.

\begin{table}[h]
\centering
\small
\caption[]{\small Texture as OOD data. We use ResNet~\citep{he2016identity} to train on the in-distribution datasets.  All values are percentages. $\uparrow$ indicates larger values are better, and $\downarrow$ indicates smaller values are better. \textbf{Bold} numbers are superior results.
}
\begin{tabular}{lrrr}
\toprule
 $\mathcal{D}_{\text{in}}$ & MS-COCO & PASCAL & NUS-WIDE  \\
 & \multicolumn{3}{c}{{ \textbf{FPR95} / \textbf{AUROC}}  / \textbf{AUPR}} \\
\textbf{OOD Score}  & \multicolumn{3}{c}{{ $\downarrow ~~~~~~~ \uparrow~~~~\uparrow$}} \\

\midrule
{MaxLogit}~\cite{hendrycks2019benchmark}
& 14.63 / 96.10 / 99.32 & 12.36 / 96.22 / 96.97
& 38.46 / 87.42 / 97.19\\
% \textbf{Prob} & Max 
% & 14.63 / 96.10 / 99.32 & 12.36 / 96.22 / 96.97
% & 38.46 / 87.42 / 97.19\\
{MSP}~\cite{hendrycks2016baseline} 
& 60.82 / 83.70 / 97.05 & 41.81 / 89.76 / 93.00
& 83.09 / 63.41 / 92.48\\
{ODIN}~\cite{liang2018enhancing}
& \textbf{12.22} / 96.18 / 99.29 &12.36 / 96.22 / 96.97   
& 38.46 / 87.42 / 97.19\\
{Mahalanobis}~\cite{lee2018simple}
& 44.61 / 85.71 / 97.41 & 19.17 / 96.23 / \textbf{97.90}
& 36.19 / 91.36 / 98.52\\
{LOF}~\cite{breunig2000lof}
& 70.16 / 74.73 / 94.96 & 89.49 / 60.37 / 76.70
& 64.27 / 78.23 / 95.94\\
{Isolation Forest}\cite{liu2008isolation}
& 95.55 / 53.21 / 90.45 &99.59 / 20.89 / 50.11
& 93.07 / 51.01 / 89.17\\
\midrule
%  \multirow{2}{0.12\linewidth}{{\textbf{Energy (ours)}}}
%   &  Max 
%   & 14.63 / 96.10 / 99.32 & 12.36 / 96.22 / 96.97
%   & 38.46 / 87.42 / 97.19\\
%  & ExpSum & 13.97 / 96.14 / 99.33 & 11.77 / 96.37 / 97.06 & 33.10 / 89.64 / 97.63 \\
\textbf{JointEnergy} (ours)  & 12.82 / \textbf{96.84} / \textbf{99.54}  & \textbf{10.87} / \textbf{96.78} / 97.87 & \textbf{31.68} / \textbf{92.43} / \textbf{98.65}\\
\bottomrule
\end{tabular}
% \vspace{-0.2cm}
% \caption[]{\small Texture as OOD data. We use ResNet~\citep{he2016identity} to train on the in-distribution datasets.  All values are percentages. $\uparrow$ indicates larger values are better, and $\downarrow$ indicates smaller values are better. \textbf{Bold} numbers are superior results.
% }
\label{tab:results-texture}
\end{table}

\begin{table}[h]
\centering
\small
\caption[]{\small Ablation study on the effect of aggregation methods for prior approaches. We use ResNet~\citep{he2016identity} to train on the in-distribution datasets. We use Texture~\citep{cimpoi14describing} as OOD test data as described in Section~\ref{sec:setup}. Note that \emph{Sum} is not applicable to tree-based or KNN-based approaches (\emph{e.g.}, LOF and Isolation Forest). 
}
\begin{tabular}{llrrr}
\toprule
&  $\mathcal{D}_{\text{in}}$ & MS-COCO & PASCAL & NUS-WIDE   \\
&  & \multicolumn{3}{c}{{ \textbf{FPR95} / \textbf{AUROC} / \textbf{AUPR}}} \\
\textbf{OOD Score} & \textbf{Aggregation} & \multicolumn{3}{c}{{ $\downarrow$ ~~~~~~~ $\uparrow$~~~~~~~~~~$\uparrow$}}\\
\midrule
\textbf{Logit} & Sum& 95.63 / 53.52 / 73.25 &96.36 / 49.44 / 43.07& {92.38 / 52.72 / 89.21}\\
\textbf{Prob} & Sum& {43.69 / 87.21 / 93.14} & {35.97 / 84.68 / 76.61} & 34.88 / 90.76 / 98.57\\
% \textbf{Logit} & Cond. Sum\\
\textbf{ODIN} & Sum & {43.69 / 87.21 / 93.14} & 53.77 / 74.50 / 67.15 & 35.27 / 89.36 / 98.31\\
\textbf{Mahalanobis} & Sum & 45.62 / 84.34 / 97.02 
& 19.45 / 96.09 / 97.80
& 37.55 / 91.04 / 98.47 \\
\textbf{LOF} & Sum & N/A & N/A & N/A \\
\textbf{Isolation Forest} & Sum & N/A & N/A & N/A \\
\midrule
\textbf{Energy} &  Sum  & \textbf{12.82} / \textbf{96.84} / \textbf{99.54}  & \textbf{10.87} / \textbf{96.78} / \textbf{97.87} & \textbf{31.68} / \textbf{92.43} / \textbf{98.65}\\
% \textbf{ODIN} & Cond. Sum \\
\bottomrule
\end{tabular}
% \vspace{-0.2cm}
% \caption[]{\small Ablation study on the effect of aggregation methods for prior approaches. We use ResNet~\citep{he2016identity} to train on the in-distribution datasets. We use Texture~\citep{cimpoi14describing} as OOD test data as described in Section~\ref{sec:setup}. Note that \emph{Sum} is not applicable to tree-based or KNN-based approaches (e.g., LOF and Isolation Forest). 
% }
\label{tab:ablation-texture}
\end{table}

\section{Baseline Methods}
In multi-label classification, the prediction for each label $y_i$ with $i \in \{1,2,...,K\}$ is independently made by a binary logistic classifier:
\begin{equation*}
p(y_i \mid \*x) = \frac{e^{f_{y_i}(\*x)}}{1+ e^{f_{y_i}(\*x)}}.
\end{equation*}
We consider the following baselines methods under \emph{maximum} aggregation:
\begin{align}
\textbf{MaxLogit}&=\max_i f_{y_i}(\*x)\\
%  \textbf{MaxProb}&=\max_i ~p(y_i \mid \*x) = \max_i \frac{e^{f_{y_i}(\*x)}}{1+ e^{f_{y_i}(\*x)}}\\
 \textbf{MSP} &=\max_i \frac{e^{f_{y_i}({\*x})}}{\sum^K_j e^{f_{y_j}({\*x})}}\\
  \textbf{ODIN} &=\max_i \frac{e^{f_{y_i}(\hat{\*x})/T}}{1+e^{f_{y_i}(\hat{\*x})/T}}\\
  \textbf{Mahalanobis}&= \max_i -(\phi(\hat{\*x}) - \hat{\mu}_{y_i})^\mathsf{T} \hat{\Sigma}^{-1} (\phi(\hat{\*x}) - \hat{\mu}_{y_i})
\end{align}
In particular, ODIN was originally designed for multi-class but we adapt for the multi-label case by taking the maximum of calibrated label-wise predictions. The input perturbation is calculated using $\hat{\*x} = \*x - \epsilon \text{sign}(-\nabla \ell_{{\hat y}_i})$, where $\ell_{{\hat y}_i}$ is the binary cross-entropy loss for the label $\hat{y}_i$ with the largest output, i.e., $\hat{y}_i = \arg \max_i p(y_i = 1 \mid \*x )$. For Mahalanobis distance,we extract the feature embedding $\phi(\*x)$ for a given sample.  $\hat{\mu}_{y_i}$ is the class conditional mean for label $y_i$, and $\hat{\Sigma}^{-1}$ is the covariant matrix. 
% $$

\subsection{Validation data for baselines} We use a combination of the following validation datasets to select hyperparameters for ODIN \citep{liang2018enhancing} and Mahalanobis \citep{lee2018simple}. The validation set consists of: 
\begin{itemize}
    \item Gaussian noise sampled i.i.d. from an isotropic Gaussian distribution;
    \item uniform noise where each pixel is sampled from $U= [-1, 1]$;
    \item In-distribution data corrupted into OOD data by applying (1) pixel-wise arithmetic mean of random pair of in-distribution images; (2) geometric mean of random pair of in-distribution images; and (3) randomly permuting 16 equally sized patches of an in-distribution image.
\end{itemize}

% ORIGINAL VERSION:
% The first anomalies are \emph{Gaussion noise} anomalies which have each dimension i.i.d. sampled from an isotropic Gaussian distribution. 
% The second ones are \emph{uniform noise} anomalies where each pixel is sampled from $U= [-1, 1]$ depending on the input space of the classifier. 
% The remaining validation sources are generated by corrupting in-distribution data, so that the data becomes out-of-distribution. 
% One such source of anomalies is created by taking the pixelwise \emph{arithmetic mean} of a random pair of in-distribution images. 
% Other anomalies are created by taking the \emph{geometric mean} of a random pair of in-distribution
% images. 
% \emph{Jigsaw anomalies} are created by taking an in-distribution example, partitioning the image into 16 equally sized patches, and permuting those patches.

\subsection{Hyperparameter tuning for baselines}
 ODIN~\citep{liang2018enhancing} and Mahalanobis~\citep{lee2018simple} require hyper-parameter tuning, such as temperature and magnitude of noise $\epsilon$. We use the validation data above for selecting the optimal hyperparameters.
For ODIN, temperature T is chosen from [1,10,100,1000] and the perturbation magnitude $\epsilon$ is chosen from 21 evenly spaced numbers starting from 0 and ending at 0.004. 
For Mahalanobis, the perturbation magnitude $\epsilon$ is chosen from [0, 0.0005,  0.0014, 0.001, 0.002, 0.005]. 
The optimal parameters are chosen to minimize the FPR at TPR95 on the validation set.

\section{Ablation Study: JointEnergy with Top Labels}
Our method can generalize to the case when using only the top-$k$ predictions in the extreme multi-label classification case, \emph{i.e.}, when the number of labels is very large. This in theory holds true when the some labels' outputs are relatively small especially when the logits are negative, and hence the label-wise energy $E_{y_i} = -\log (1+e^{f_{y_i}(\mathbf{x})}) \approx 0$. In this case, omitting these labels does not largely affect the overall JointEnergy score and detection performance. The results are shown in Table ~\ref{tab:topk}. 

\begin{table}[h]
\centering
\small
\caption[]{\small Ablation study on JointEnergy with top-$k$ labels. $k$ is the average number of labels of a training image (estimated empirically on the entire training population) for each dataset. Specifically, $k$ is 3 for MS-COCO, 2 for PASCAL-VOC, and 2 for NUS-WIDE. Dataset and model settings are the same as in Table ~\ref{tab:results-imagenet}.
}
\begin{tabular}{lrrr}
\toprule
 $\mathcal{D}_{\text{in}}$ & MS-COCO & PASCAL & NUS-WIDE  \\
 & \multicolumn{3}{c}{{ \textbf{FPR95} / \textbf{AUROC}}  / \textbf{AUPR}} \\
\textbf{OOD Score}  & \multicolumn{3}{c}{{ $\downarrow ~~~~~~~ \uparrow~~~~\uparrow$}} \\

\midrule
{JointEnergy (all)}   & \textbf{33.48} / \textbf{92.70} / \textbf{96.25}  & \textbf{41.01} / \textbf{91.10} / \textbf{86.33} & \textbf{48.98} / \textbf{88.30} / \textbf{96.40}\\
{JointEnergy (top-$k$)} &  37.85 / 91.69 / 95.71
& 43.63 / 90.39 / 85.35
& 53.00 / 86.35 / 95.68 \\
{JointEnergy (top-5)} & 35.83 / 92.19 / 95.99
& 41.36 / 91.00 / 86.22
& 49.87 / 87.94 / 96.28 \\
{JointEnergy (top-10)} & 34.43 / 92.52 / 96.16
& 41.02 / 91.09 / 86.32
& 49.35 / 88.22 / 96.37 \\
{JointEnergy (top-20)} & 33.72 / 92.66 / 96.23
& 41.01 / 91.10 / 86.33
& 49.14 / 88.28 / 96.39 \\
{JointEnergy (top-40)} & 33.53 / 92.70 / 96.25
& - / - / -
& 48.98 / 88.30 / 96.40 \\
{JointEnergy (top-60)} & 33.48 / 92.70 / 96.25
& - / - / -
& 49.03 / 88.30 / 96.40 \\
% \textbf{ODIN} & Cond. Sum \\
\bottomrule
\end{tabular}
% \vspace{-0.2cm}
% \caption[]{\small Ablation study on JointEnergy and top-k. k is the average number of labels of a training image (estimated empirically on the entire training population) for each dataset. Specifically, k is 3 for MS-COCO, 2 for PASCAL-VOC, and 2 for NUS-WIDE. Dataset and model settings are the same as in Table ~\ref{tab:results-imagenet}.
% }
\label{tab:topk}
\end{table}

% l_j(\*x) &= y_{\hat{i}} \cdot \log(p(y_{\hat{i}} = 1 | \*x)) + (1 - y_{\hat{i}}) \cdot \log(p(y_{\hat{i}} = 0 | \*x))
% $$

% For ODIN, we also consider different ways to compute the binary cross-energy loss w.r.t the aggregation. For max, we compute the loss for each class and pick the one with highest probability. For sum, we compute the loss for each class and average across labels.
% \begin{align}
% \textbf{Max}:~~ l_j(\*x) &= y_{\hat{i}} \cdot \log(p(y_{\hat{i}} = 1 | \*x)) + (1 - y_{\hat{i}}) \cdot \log(p(y_{\hat{i}} = 0 | \*x))\\
% where \; \hat{i} &= arg \max_i p(y_i = 1 | \*x )\\[5pt]
%  \textbf{Sum}:~~ l_j(\*x) &= \sum_{i=1}^{K} y_i \cdot \log(p(y_i) = 1 | \*x) + (1 - y_i) \cdot \log(p(y_i) = 0 | \*x)\\
% \end{align}

% After getting the loss, we compute the gradient and add the perturbations. 
% \begin{align}
% \hat{\textbf{x}} &= \textbf{x} - \epsilon sign(-\nabla\frac{1}{N}\sum_{j}^{N}l_j) \\ 
% S(\hat{\textbf{x}};T) &= \max_i \frac{e^{f_i(\hat{\textbf{x}})/T}}{1+e^{f_i(\hat{\textbf{x}})/T}}
% \end{align}

\end{document}